\documentclass{article}

\usepackage{arxiv}

\usepackage{hyperref}       
\usepackage{url}  
\usepackage{microtype}
\usepackage{graphicx}
\usepackage{subcaption}

\usepackage{multirow}
\usepackage{colortbl}
\usepackage{mdframed}
\usepackage{booktabs} 
\usepackage[table,xcdraw]{xcolor} 

\usepackage{amsmath}
\usepackage{amssymb}
\usepackage{mathtools}

\title{Energy-Conscious LLM Decoding:  \\
Impact of Text Generation Strategies on GPU Energy Consumption}

\author{
 Alireza Nik \\
  SimulaMet\\
  Oslo, Norway \\
   \And
 Michael A. Riegler \\
  Simula\\
  Oslo, Norway \\
  \And
 Pål Halvorsen \\
  SimulaMet\\
  Oslo, Norway \\
}

\begin{document}
\maketitle
\begin{abstract}
Decoding strategies significantly influence the quality and diversity of the generated text in Large Language Models (LLMs), yet their impact on computational resources, particularly GPU energy consumption, is insufficiently studied. This paper investigates the relationship between text generation decoding techniques and energy efficiency, focusing on the trade-off between generation quality and GPU energy usage across diverse tasks and decoding configurations. By benchmarking multiple strategies across various tasks, including Translation, Math Problem Solving, Coding, and Open-ended text generation, we reveal how selecting appropriate decoding techniques with their tuned hyperparameters affects text quality and has measurable implications for energy consumption. Our findings show that the choice of decoding strategy can greatly impact GPU energy usage, even when it has a minimal effect on output quality. Different strategies also involve trade-offs between quality and energy efficiency, and no single decoding method is best in all cases across every metric. To the best of our knowledge, this is one of the first studies to examine decoding strategies in LLMs from the perspective of energy consumption, providing useful insights for building energy-efficient applications without compromising text generation quality.
\end{abstract}

\section{Introduction}
\label{intro}

In the past few years, we have witnessed a significant increase in the applications of Large Language Models (LLMs). Powerful text generation LLMs such as OpenAI's ChatGPT \cite{chatgpt}, Google’s Gemini \cite{gemini}, and Anthropic’s Claude Sonnet \cite{sonnet} have been completely intertwined in our daily lives. The popularity and increasing applications of LLMs across various domains has caused the AI community to continuously scale the model sizes up to hundreds of billion parameters \cite{palm}. This growth has led to a significant surge in computational demands causing a considerable energy consumption associated with the hardware these models are being deployed, especially the GPUs \cite{nine,ten}. GPUs have become indispensable to modern high-performance computing (HPC) infrastructure due to their efficient parallel processing \cite{sixteen}. This enhanced capability has made them an excellent accelerator for training and deploying LLMs. However, their significant energy consumption has raised serious alarms recently due to their increasing scale and the global push towards sustainable AI \cite{nine,twelve}. Although most of the AI environmental research have revolved around the energy consumption and carbon footprints during training LLMs \cite{thirteen,patterson2021carbon,fourteen,fifteen,seventeen}, the energy concerns associated with LLM inference have been studied much less. Yet, the LLM inference demand is substantially larger \cite{patterson2022carbon,patel2024characterizing}. Among recent studies focusing on the energy footprint of LLM inference is an analysis by \cite{epochAI}, which measured the energy consumption of GPT-4o single query. Their findings show that earlier research presented by \cite{de2023growing} estimating ChatGPT’s energy usage at about 3 watt-hours per query was significantly exaggerated, with the more accurate estimation being closer to 0.3 watt-hours per query. Similarly, a recent study by \cite{elsworth2025measuring} reported an energy consumption of 0.24 watt-hours per prompt in Google’s Gemini. Similarly, \cite{nine} reported the energy usage of different LLaMA model sizes. For the LLaMA-65B model, they reported that with a maximum generation length of 512 tokens, the energy consumption per query is approximately 0.3 watt-hours.

On the other hand, the ability of LLMs to generate high-quality coherent response text is significantly influenced by the choice of decoding strategies during inference. Decoding strategies transform the probability distributions generated by LLMs into fluent and relevant output text \cite{one,two,three}. They determine how tokens are selected at each step of text generation. This transformation termed as decoding is as crucial as the model itself in achieving the best quality of the generated text. However, the selection of the appropriate decoding strategy and tuning their specific parameters remain insufficiently explored \cite{one}. In most cases, the evaluations of the newly proposed, state-of-the-art LLMs are only conducted using the default decoding method, instead of studying the effects of adopting different strategies and varying their parameters to achieve the best benchmarking results. Prior research has primarily focused on the introduction of new text generation decoding strategies in the past few years \cite{four,five,six,seven,eight}. Each one of these techniques claims to be a superior method based on text quality metrics on specific text generation tasks.  While the effects of various decoding strategies on text quality, diversity, and coherence have been studied recently \cite{one,two}, their implications for computational efficiency and energy consumption remain unexplored. Decoding strategies differ in their computational complexity and, consequently, their energy demands. For instance, beam search, which explores multiple hypotheses simultaneously, is more computationally demanding compared to simpler methods like greedy search. Understanding how these decoding techniques affect energy consumption is necessary for optimizing the LLM inference in resource-constrained devices and achieving an appropriate balance between model performance and sustainability goals. 
In this respect, the main contribution of this research is to analyze the impact of decoding strategies on GPU energy consumption, including:

\begin{itemize}\itemsep0em 
\item \textit{Measuring and comparing GPU energy consumption across various decoding strategies}
\item \textit{Analyzing the trade-offs between energy efficiency and generated text quality for different decoding techniques}
\item \textit{Investigating the relationship between hyperparameter configurations and GPU energy consumption across decoding methods}
\item \textit{Developing an evaluation framework combining energy monitoring, quality metrics and statistical analysis across multiple NLP tasks}
\item \textit{Introducing new efficiency metrics that combine generation quality and energy consumption to enable systematic comparison of decoding strategies}
\item \textit{Quantifying the stability and robustness of different decoding methods through hyperparameter sensitivity analysis}
\item \textit{Distinguishing between decoding latency and GPU energy consumption by analyzing their relationship}
\end{itemize}

Through these contributions, our work adds to the growing literature on the generative performance and energy characteristics of LLM inference. We also hope it motivates further research, benchmarking, and open sharing of systematic performance insights across a broader range of models and hardware settings. Such efforts are particularly valuable given the growing concerns about the energy consumption and environmental impact of large-scale AI models, emphasizing the need for developing more efficient inference methods.

\section{Background}
\label{background}

LLMs generate text sequentially in an autoregressive manner. At each step, the model computes a probability distribution over the vocabulary. Decoding strategies determine how the model converts these probabilities into generated text. The choice of decoding method significantly influences the quality, diversity, and coherence of the output. These strategies are generally categorized into two types: deterministic and stochastic.

\subsection*{Deterministic Strategies}
\label{deterministic decoding}

Deterministic decoding prioritizes maximizing the probability of output sequences according to the model’s learned distribution \cite{six}. Although these strategies lead to higher generative accuracy and coherence, they also limit diversity and creativity, which can result in text degeneration \cite{topk,topp}. In this study, we examine a selection of widely adopted deterministic methods:

\begin{itemize}\itemsep0em 
    \item \textbf{Greedy:} The most basic decoding strategy where the model select the token with the highest probability at each generation step \cite{two,consistency}. Since tokens with the highest probability are always chosen while the rest are completely ignored, this technique may leads to repetition and degeneration, especially in tasks where diversity in the output text is an important factor.
    
    \item \textbf{Beam Search:} Instead of greedily selecting the tokens, Freitag et al. \cite{beam} proposed keeping track of multiple candidate sequences (a beam), expanding them at each generation step, and selecting the top ones based on their overall scores after multiple steps. The number of these candidate sequences is controlled by the beam width, which serves as a hyperparameter.
    
    \item \textbf{Diverse Beam Search (DBS):} Vijayakumar et al. \cite{dbs} introduced an extension of the original Beam Search method by dividing the candidate sequences into multiple subgroups while enforcing diversity within each group. The key hyperparameters for this approach are the beam width and the number of groups.
    
    \item \textbf{Contrastive Search (CS):} Su et al. \cite{six} introduced a contrastive framework to penalize repetitive token selections while maintaining coherence. After selecting the top-k candidates from the model, they are assessed based on the likelihood assigned by the LLM and a degeneration penalty that evaluates how close the token is relative to the previous context. This method have two hyperparameters: k, which controls candidate selection based on top-k, and \( \alpha\), which controls degeneration.
    
    \item \textbf{DoLa:} Chuang et al. \cite{seven} designed a decoding technique to improve factual accuracy in LLMs. Unlike other decoding methods that rely solely on the final layer’s logits, DoLa compares the output logits of earlier layers with those of the final layer. The specific layers to be contrasted with the final one can be set as a hyperparameter.
    
    \item \textbf{Assisted Decoding (AD):} A novel strategy, also known as speculative decoding, was introduced by Leviathan et al. \cite{speculative} to accelerate LLM inference. This technique uses an assistant model to generate multiple candidate tokens in advance, which are then verified by a larger target model. In this way, the number of sequential computations during inference is significantly reduced, speeding up the decoding process. Since we used only the n-gram-based speculative decoding in the greedy setting and not the model-based approach, we categorize it as deterministic. The hyperparameter in this case is the number of tokens to be output as candidate tokens.
\end{itemize}

\subsection*{Stochastic Strategies}
\label{stochastic decoding}

Stochastic decoding introduces randomness in the text generation process to help prevent text degeneration. These methods increase the model’s ability to generate diverse outputs, which is especially useful for creative writing. While these sampling-based methods help increase diversity, the randomness in these approaches can negatively affect the coherence of the generated text \cite{six}. This work considers the following stochastic decoding methods for evaluation:

\begin{itemize}\itemsep0em 
    \item \textbf{Temperature (\( \tau\)):} This sampling method simply modifies the logits before applying the Softmax \cite{bridle} function in the decoding process. When the temperature is low, the method becomes more deterministic by increasing the likelihood of selecting high-probability tokens. In contrast, selecting higher temperature values introduce randomness in generation by increasing the likelihood of choosing lower-probability tokens.
    
    \item \textbf{Top-k:} Fan et al. \cite{topk} introduced a method which sampling is performed from the \( k\) most probable candidates in the vocabulary at each token generation step. A smaller \( k\) makes the output more deterministic, whereas a larger k allows tokens with lower probabilities to be considered in the selection process.
    
    \item \textbf{Top-p (Nucleus):} Rather than selecting a fixed number of tokens and sampling from them, Holtzman et al. \cite{topp} proposed selecting tokens based on their probability distribution. They suggested choosing the minimal set of tokens whose cumulative probability goes beyond a probability \( p\)
    
    \item \textbf{Epsilon Sampling:} Hewitt et al. \cite{epsilon} introduced another truncation sampling approach by selecting only the tokens whose conditional likelihoods are above an entropy-dependent threshold \( \epsilon\). This ensures that the highly unlikely tokens are excluded. 
    
    \item \textbf{Typical Sampling:} To generate text that resembles most to human writing, Meister et al. \cite{typical} proposed selecting tokens based on information entropy. In the sampling approach, we select tokens based on how close a token’s probability is to the expected information content at each step. Thus, tokens that are too predictable or too unlikely are discarded, with the selection process controlled by a typicality threshold.
    
    \item \textbf{Min-p Sampling:} Nguyen et al. \cite{eight} proposed a dynamic token selection strategy to overcome the limitations of other truncation-based sampling methods. Their approach adjusts the filtering threshold at each step based on the probability of the most likely token. The key hyperparameter is a base probability, which is multiplied by the likelihood of the most probable token to dynamically adjust the threshold for the next generation step. 
\end{itemize}

\section{Experimental Setting}
\label{experiments} 

\subsection{Compute Resources} 

All the experiments were conducted on a compute node equipped with two NVIDIA A100-PCIE-40GB GPUs and a Dual 64-core AMD EPYC 7763 Processor @ 2.45GHz. The GPUs have the Thermal Design Power (TDP) of 250 watts. The system is managed by SLURM \cite{slurm}, an open-source workload manager that facilitates resource allocation, job scheduling, and queue management. In each job (experiment), we requested only one GPU. Before running the main job, we start monitoring and collecting the GPU resource utilization data over the job’s run time. This monitoring is stopped immediately after the main job is finished.

All the experiments were run on an isolated compute node to consider the variations in the LLM inference measurements and to minimize the external interference. When running the same experiment (a specific decoding method with fixed parameters with same input data) multiple times alongside other users’ jobs in a compute node, we sometimes observed significant variations in the total inference time and consumed energy. These fluctuations in LLM inference workloads are due to the shared nature of the system (the potential disruptions of other jobs) and distinct compute characteristics of LLMs \cite{eleven}. Consequently, we submitted all our jobs on isolated node to mitigate the significant variations. 

\subsection{Monitoring Tools}

We use Nvidia built-in power sensors to measure GPU power consumption in our experiments. Nvidia offers two on-board monitoring tools: Nvidia System Management Interface\cite{smi}(nvidia-smi) and Nvidia Data Center GPU Manager\cite{dcgmi} (dcgm). Nvidia-smi is a command-line utility to monitor and manage Nvidia GPUs. It provides detailed information about GPU performance metrics including utilization, temperature, memory usage, memory clock frequency, power consumption, and etc. While this utility is useful for monitoring, it offers only a limited set of metrics \cite{profiling}. In contrast, dcgm provides more fine-grained insights than nvidia-smi. It can capture detailed hardware utilization and an easier GPU management in data centers. Both tools allow us to record power consumption at a sampling frequency in units of milliseconds.

\subsubsection{Energy Measurement Procedure}

We conducted a thorough investigation of what granularity of power measurements works best in our experiments. According to the dcgm documentation, the minimum sampling frequency supported is 100 ms, but the nvidia-smi documentation does not provide any information about its minimum sampling interval. We tested sampling frequencies of 10, 100, and 1000 ms with nvidia-smi. While finer-grained measurements would be ideal, we observed important technical limitations when using on-board monitoring tools. With nvidia-smi, shorter sampling intervals led to significant information loss compared to the 1000 ms interval. For example, at a requested 100 ms interval, we collected only about 9 samples per second instead of the expected 10, indicating a 10\% information loss. This issue was even more pronounced at 10 ms, where we gathered only about 70 samples per second instead of 100, representing a 30\% loss. These results show that nvidia-smi is less reliable for power monitoring at small sampling intervals. These technical limitations of nvidia-smi, due to its internal mechanisms, have also been reported by Yang et al. \cite{yang2024accurate} and Zou et al. \cite{indicator}. 

We also tested sampling intervals of 1000 and 100 ms with the NVIDIA dcgm tool. At the minimum frequency of 100 ms, we observed much less sampling loss compared to the same case with nvidia-smi, showing that dcgm is a better monitoring tool for collecting fine-grained power measurements. When comparing the aggregated energy consumption across different decoding strategies on the watt-hour scale, we found only small differences between nvidia-smi (with a 1000 ms interval) and dcgm (with a 100 ms interval). Importantly, the relative rankings of decoding strategies remained consistent across both tools. However, to better capture momentary power draw fluctuations during LLM inference and instantaneous spikes in GPU utilization, we collected fine-grained power measurements using the dcgm tool with a 100 ms sampling interval. 

To quantify the total energy consumption, we rely on the \( E = \sum_{i=1}^{n} P_i \cdot \Delta t_i\) formula that relates energy to power (\( P_i\)) and time period (\(\Delta t_i\)). Since we recorded power measurements at 100-millisecond intervals (\(\Delta t_i = 0.1s\)), assuming constant power (in watts) over each sampling period, the total energy consumption in joules is equivalent to the summation of the recorded power values over the entire inference duration multiply 0.1. GPU energy consumption, especially in LLM workloads, results in significantly large values when measured in joules, as joules are a relatively small unit for quantifying energy. To facilitate easier comparison between decoding methods, we converted the energy consumption from joules (J) to watt-hours (Wh) for reporting in this  (\( 1\mathrm{Wh} = 3.6 \times 10^3 \, \mathrm{J}\)).

\subsection{Benchmarks}

For reproducibility, we used LM-evaluation-harness \cite{eval-harness} to benchmark and evaluate the chosen LLMs across different decoding methods. This is an open-source framework designed for evaluating and testing the performance of the LLMs across a wide range of tasks and benchmarks. We evaluated our selected decoding strategies across various text generation tasks: \textit{Translation}, \textit{Coding}, \textit{Math}, and \textit{Open-ended Generation}. For all tasks, we utilized only a subset of the full datasets. Using the complete dataset would unnecessarily increase inference time and energy consumption during experimentation, without bringing any meaningful insights to the study. We emphasize that we reported only the available performance metrics for the generated texts as provided by the LM-evaluation-harness framework (v0.4.8) across our selected tasks.   

\subsubsection{Translation}

We selected the WMT16 \cite{wmt2016} dataset to evaluate the translation performance. WMT is a widely recognized competition for machine translation tasks, providing high-quality datasets across multiple languages. In our experiments, we focused on German-English and English-German and only considered the first 500 input prompts of the whole dataset in both cases. For evaluation, We reported the \textit{BLEU} \cite{bleu} metric. This widely-used evaluation metric measures the n-gram overlap between the generated output and the ground truth, with higher scores indicating the better generative performance.

\subsubsection{Math} 

We utilized the GSM8K \cite{gsm8k} dataset to evaluate the reasoning and mathematical problem-solving capabilities of LLMs under different decoding strategies. Although this popular dataset consists of 8K grade-school-level math problems, we only focused on the first 100 problems out of the whole benchmark. In this specific task, we used 5-shot prompts in our experiments.  For this benchmark, the \textit{Exact Match} accuracy is reported as the performance of the generated text compared to the ground truth. This metric checks if each token in the a language model's output aligns with the greedy decoding generation of the input at each step \cite{lessons}. If all tokens align perfectly, the LLM's generated text is considered an exact match. Thus, a higher value indicates better text generation performance.

\subsubsection{Coding}

The HumanEval \cite{chen2021evaluating} benchmark suit is a collection of programming problems designed to evaluate the AI models on code generation capabilities. The benchmark asses an LLM's ability in code synthesis tasks, typically measured using the pass@k metric. We evaluated our models on 100 programming problems in the benchmark and reported the pass@1 accuracy.

\subsubsection{Open-ended Generation}

We also evaluated our models on the WikiText \cite{merity} benchmark, a large dataset of verified articles on Wikipedia curated for open-ended text generation. Wikitext is a standard benchmark for evaluating LLMs in free-form language modeling. We utilized 400 randomly selected articles from the pre-processed dataset provided by Su et al. \cite{su2022empirical} in our experiments. To evaluate WikiText benchmark, we used MAUVE \cite{mauve} metric. It quantifies how similar the distribution of generated text is to that of human-written reference text. The resulting score reflects the overlap between the two distributions.

\begin{table*}[b]
    \centering
    \caption{Decoding Strategies and Selected Hyperparameters}
    \begin{small}
    \begin{sc}
    \begin{tabular}{lccr}
    \toprule
    \textbf{strategies} & \textbf{hyperparameter value} \\
    \midrule
    min\_p & {\{0.02, 0.05, 0.1, 0.3, 0.5, 0.7, 0.9, 1.0\}}\\
    epsilon & {\{0.0003,  0.0006,  0.0009,  0.002, 0.004\}}\\ 
    contrastive & {\{\(\alpha:\){\{0.2, 0.4, 0.6\}}  \(k:\){\{5, 25\}}\}}\\ 
    temperature & {\{0.1, 0.3, 0.5, 0.7, 0.9, 1.2\}}\\ 
    diverse beam & {\{(4,2), (5,5), (10,2), (10,5)\}}\\ 
    top\_p & {\{0.3, 0.5, 0.7, 0.8, 0.95\}}\\ 
    top\_k & {\{5, 25, 50, 100\}}\\
    typical & {\{0.5, 0.8, 0.9, 0.95\}}\\
    dola & {\{low, high\}}\\ 
    assisted & {\{2, 5, 10\}}\\ 
    beam search & {\{2, 5, 10\}}\\
    \bottomrule
    \end{tabular}
    \end{sc}
    \end{small}
    \vskip -0.1in
    \label{table-hyper}
\end{table*}

\subsection*{LLMs} 

In our experiments, we used Qwen 2.5 \cite{qwen2.5} and LLaMA 3.1 \cite{llamaPaper}, two state-of-the-art family of open-source models. Specifically, we experimented with the 7B and 14B instruction-tuned variants of Qwen 2.5, as well as the 8B instruction-tuned variant of LLaMA 3.1. By using models of different sizes and from different model families, our goal is to analyze how architectural choices and parameter scaling influence decoding behavior, runtime, and energy consumption in different inference tasks.    

\subsection*{Strategies \& Hyperparameters}  

We employed the decoding strategies introduced in Sections \ref{deterministic decoding} and \ref{stochastic decoding} in our study. All adopted strategies have been implemented in the Hugging Face \cite{transformers} library (v4.51.3). We also tuned the hyperparameters of each strategy to evaluate their impact on text quality and energy consumption across our selected tasks. The range of hyperparameter values for each decoding technique was selected based on recommendations from standard practices and recent studies \cite{one,two}. However, in some cases, we included unconventional hyperparameter values to investigate their impact, particularly on energy consumption. Table \ref{table-hyper} lists the utilized decoding strategies and their selected hyperparameters. Our goal is to report the optimal decoding hyperparameters for our benchmark suite across different models and to examine the relationship between performance and energy consumption in comparable settings for each decoding strategy. By focusing on a limited subset of datasets, we aim to determine which decoding strategy is more energy-efficient during LLM inference.

\subsection{Benchmarking Setup}

We conducted an extensive benchmarking of different decoding strategies across five text generation tasks. The batch size is set to 1 in all experiments to prioritize comparability. For each decoding method, we experimented multiple hyperparameters, repeating each parameter setting 5 times to ensure reliability and to capture energy variations in the LLM's inference. During each run, we evaluated text generation performance and recorded GPU energy consumption, GPU utilization and inference latency. Through preliminary testing, we found that 5 repetitions were sufficient to capture potential variations or spikiness in GPU utilization. We then reported the average energy consumption across all runs. We also ensured that the SM and memory clock frequencies of both GPUs remained consistent across all experiments, since variations in clock frequency can directly affect GPU power draw and, in turn, our energy measurements. After completing each job, We utilized a post-processing pipeline that identifies the active GPU accelerator periods. In other words, we filtered the recorded logs to focus solely on the active GPU periods. This approach allowed us to accurately quantify the GPU energy consumption and ensured that our experiments maintained consistent and stable GPU utilization throughout the generation process without any GPU idle periods.

\section{Experimental Results \& Discussions}

We examine the trade-offs between text generation quality and GPU energy consumption for each decoding strategy, analyze how hyperparameter tuning impacts energy usage and explain how energy efficiency can be different from decoding latency in the following sections.

\subsection{Text Generation Quality vs. Energy Efficiency}
\label{experimental:TG-EF}

\begin{figure*}[t]
\centering
\begin{subfigure}{0.32\textwidth}
  \includegraphics[width=\linewidth]{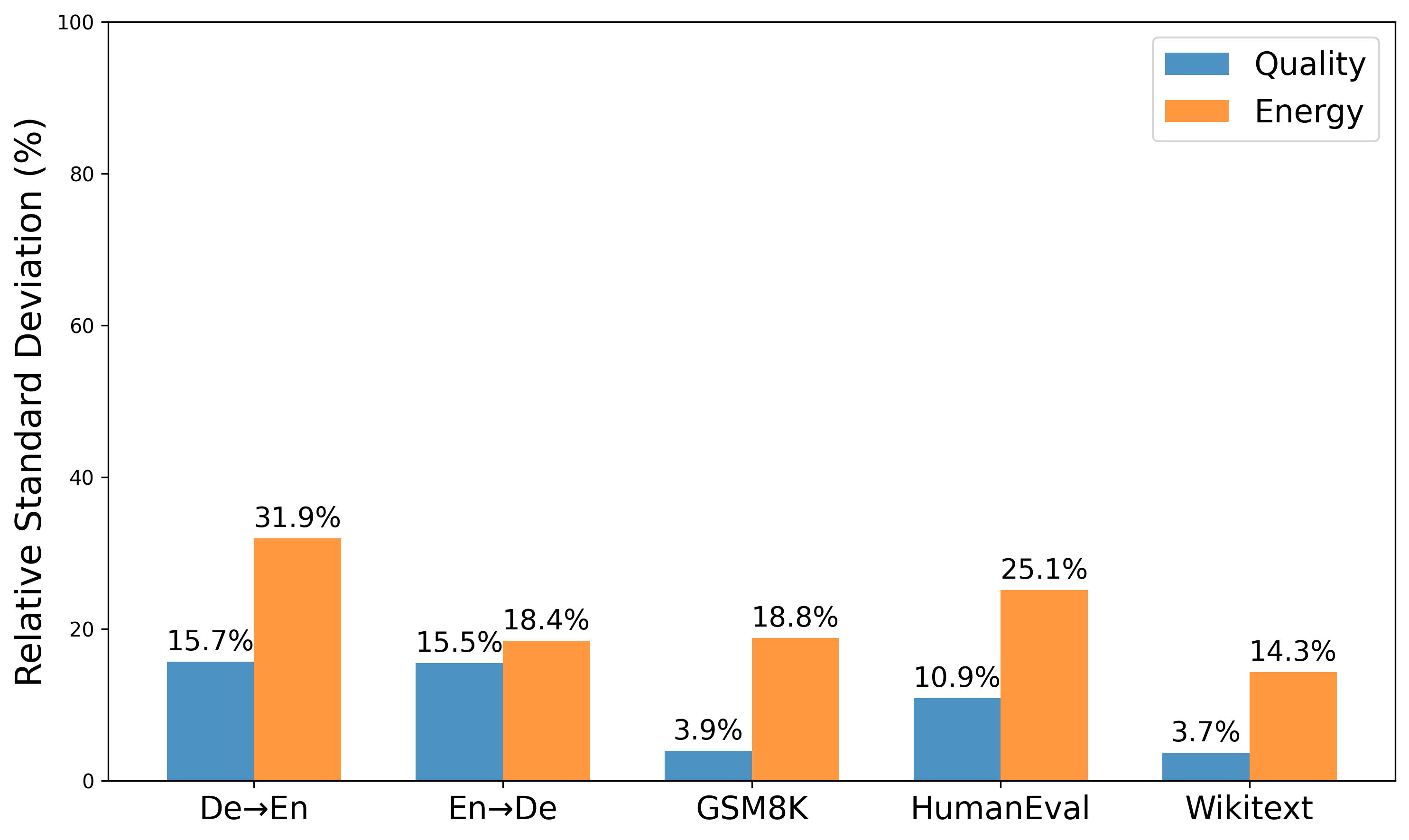}
  \caption{Qwen2.5-7B-Instruct}
  \label{rsd-dataset-qwen7B}
\end{subfigure}
\hfill
\begin{subfigure}{0.32\textwidth}
  \includegraphics[width=\linewidth]{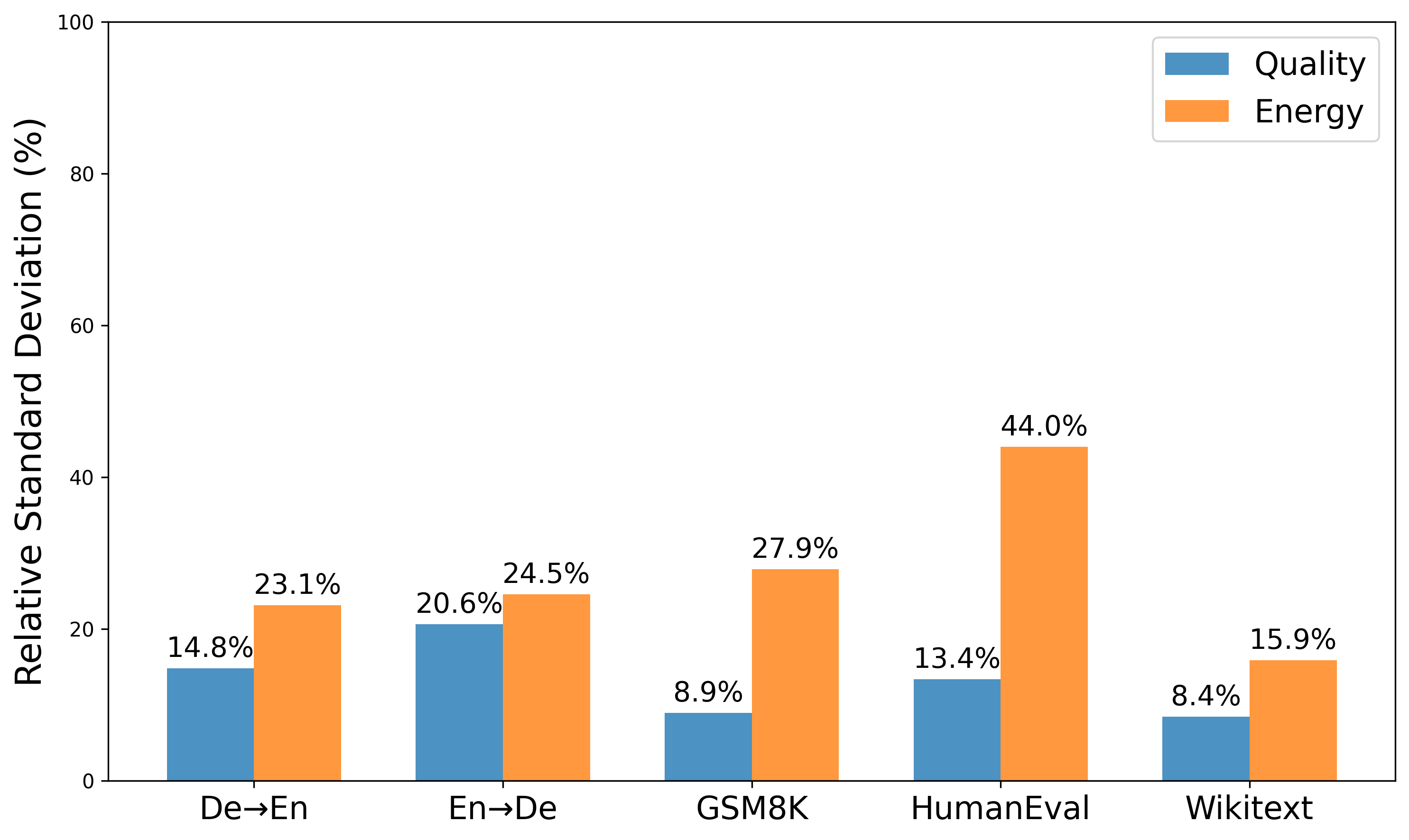}
  \caption{Llama3.1-8B-Instruct}
\end{subfigure}
\hfill
\begin{subfigure}{0.32\textwidth}
  \includegraphics[width=\linewidth]{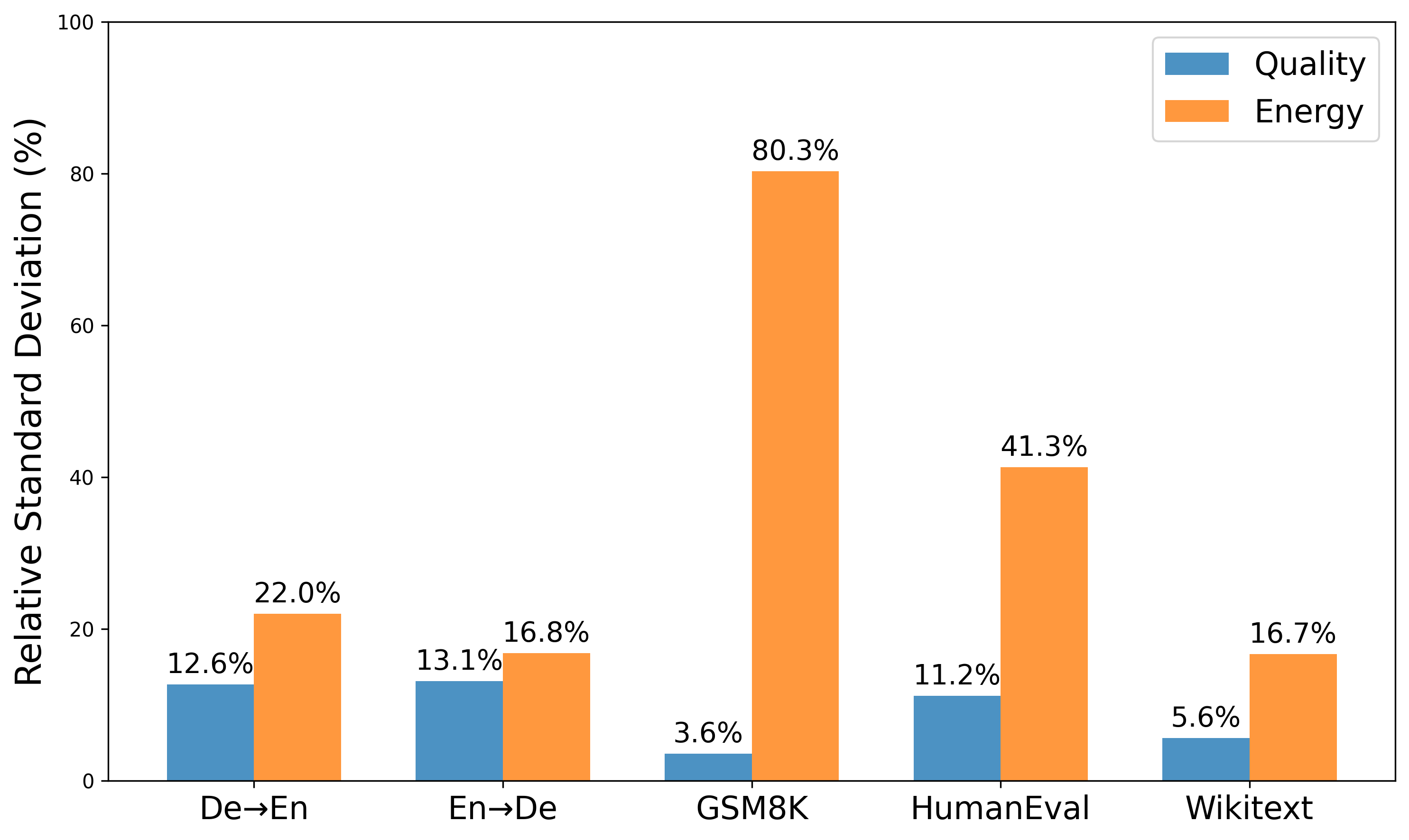}
  \caption{Qwen2.5-14B-Instruct}
\end{subfigure}

\caption{Relative Standard Deviation of text generation quality and GPU energy usage across decoding strategies.}
\label{rsd-dataset-all}
\end{figure*}

The text generation quality and energy consumption for various decoding strategies using the Qwen2.5 7B-Instruct model are presented in Table \ref{tab:performance_energy_comparison_7B}. The results were obtained using the best hyperparameter configuration that delivered the highest quality for each decoding method. The corresponding energy consumption for this optimal hyperparameter was averaged across all 5 runs and reported in watt-hours. Similar procedure is used for reporting the results of Llama3.1 8B-Instruct and Qwen2.5 14B-Instruct models in Tables \ref{tab:performance_energy_comparison_llama} and \ref{tab:performance_energy_comparison_14B} in Appendix, respectively. To demonstrate the effect of decoding strategies on generative quality and energy consumption, and to test whether the differences between strategies are statistically significant, we used the Friedman\cite{friedman} test, as our measurements did not follow a normal distribution. This non-parametric test checks for significant differences between two or more groups of repeated measurements. In our case, the repeated measurements are the quality and energy consumption of different decoding strategies in different scenarios (tasks \( \times\) models). We applied the Friedman test separately on quality and energy measurements across all models and tasks (\( n=15\) blocks, \( df=11\)). For quality measurements, we obtained \( X^2=68.77\) with \( p=2.09e^{-10}\). For energy measurements, the results were \( X^2=121.75\) with \( p=8.07e^{-21}\). In both cases, the tests were significant, showing that there are clear differences in either quality or energy consumption values of various decoding strategies across the different groups. We also ran the same statistical test for each model and each task separately across different decoding strategies for both quality and energy values. The results are listed in Tables \ref{tab:friedman_model} and \ref{tab:friedman_task} in Appendix. We observe that the tests were significant in all cases for both quality and energy values, with only one exception in the per-task analysis of quality. This confirms that the choice of decoding strategy consistently has a significant impact on both generative quality and energy consumption, regardless of the specific model or task.

We evaluated the trade-off between quality and energy across different decoding methods using an Efficiency Ratio (ER = Quality/Energy). This metric measures the generative quality achieved per unit of energy consumed, making it suitable for identifying methods that offer high quality while minimizing energy usage. However, care must be taken in interpretation, as the metric can unfairly favor cases where generative accuracy is very low compared to a baseline but energy usage is also low \cite{yang2024double}.

As shown in Table \ref{tab:performance_energy_comparison_7B}, in both translation tasks, Beam Search and DBS achieve the highest BLEU scores, up to about 42.7 for De→En and 29.9 for En→De, but also draw considerable power, pulling down their overall efficiency ratio. In contrast, AD delivers BLEU scores near—or on par with—those top‐performing beam‐based methods while consuming far less power, thereby attaining the best efficiency ratios. Greedy decoding, along with Top‐p, and Min‐p sampling methods, occupies a middle ground: they offer moderately high BLEU at moderate energy costs, resulting in fairly respectable efficiency.

Similarly, in GSM8K, Beam search achieves the highest exact-match accuracy (0.87), but it requires 14.44 Wh of energy, resulting in a modest efficiency ratio. In contrast, Top-p maintains solid accuracy (0.84) at just 12.78 Wh, yielding one of the highest efficiency ratio among all methods. Several stochastic methods also strike a better balance than beam-based approaches, generating high-quality texts at more moderate energy levels.

\begin{table*}[t]
\footnotesize
\centering
\caption{Text Generation Quality and Average Energy Consumption across Decoding Methods in their best Hyperparameter Setting using Qwen2.5-7B-Instruct with batch size=1. The associated hyperparameters are listed in Table \ref{tab:optimal_hyperparameter_7B}. The table uses color coding to highlight key metrics: blue indicates high generative quality, red signifies high energy consumption, and green represents a high efficiency ratio. $\uparrow$ indicates higher is better, $\downarrow$ indicates lower is better.}
\label{tab:performance_energy_comparison_7B}

\centering
\renewcommand{\arraystretch}{1.3} 
\resizebox{0.99\textwidth}{!}{
\setlength{\tabcolsep}{4pt} 
\begin{tabular}{!{\vrule width 1.2pt}c|c!{\vrule width 1.2pt}c|c|c|c|c|c!{\vrule width 1.2pt}c|c|c|c|c|c!{\vrule width 1.2pt}}
\noalign{\hrule height 1.2pt}
\multirow{2}{*}{\textbf{Task}} & \multirow{2}{*}{\textbf{Metric}} & \multicolumn{6}{c!{\vrule width 1.2pt}}{\textbf{Deterministic Methods}} & \multicolumn{6}{c!{\vrule width 1.2pt}}{\textbf{Stochastic Methods}} \\
\cline{3-14}
& & \textbf{Greedy} & \textbf{Beam} & \textbf{DBS} & \textbf{DoLa} & \textbf{CS} & \textbf{AD} & $\tau$ & \textbf{Top-p} & \textbf{Top-k} & \textbf{Eps} & \textbf{Typical} & \textbf{Min-p} \\
\noalign{\hrule height 1.2pt}
\multirow{3}{*}{De→En} & $BLEU \uparrow$ & 41.09 &\cellcolor{blue!35} 42.71 &\cellcolor{blue!35} 42.65 & 22.97 & 41.03 &\cellcolor{blue!30} 41.76 & 41.02 & 41.26 & 31.51 & 31.53 & 35.68 & 41.29 \\
& $Wh\downarrow$ & 9.39 & 11.13& \cellcolor{red!40} 14.23 & \cellcolor{red!50}22.29 &\cellcolor{red!40} 13.26 & 8.23 & 9.54 & 9.37 & 10.28 & 10.09 & 10.05 & 9.46 \\
\cline{2-14} 
& $ER\uparrow$ &\cellcolor{green!50} 4.37 & 3.83 & 2.99 & 1.03 & 3.09 &\cellcolor{green!50} 5.07 & 4.29 &\cellcolor{green!50} 4.40 & 3.06 & 3.12 & 3.55 &\cellcolor{green!50} 4.36\\
\hline
\multirow{3}{*}{En→De} & $BLEU \uparrow$ & 26.66 &\cellcolor{blue!35} 29.88 &\cellcolor{blue!35} 29.33 & 19.11 &\cellcolor{blue!30} 26.94 & 26.47 & 26.75 & \cellcolor{blue!30}26.96 & 19.75 & 18.12 & 22.13 & 26.85 \\
& $Wh\downarrow$ & 13.01 & \cellcolor{red!40}17.87  & \cellcolor{red!40}17.43  &\cellcolor{red!50} 21.86  & 15.40 & 12.07 & 13.14 & 13.24 & 13.42 & 13.67 & 13.54 & 13.28 \\
\cline{2-14} 
& $ER\uparrow$ &\cellcolor{green!50} 2.04 & 1.67 & 1.68 & 0.87 & 1.74 &\cellcolor{green!50} 2.19 &\cellcolor{green!50} 2.03 &\cellcolor{green!50} 2.03 & 1.47 & 1.32 & 1.63 & \cellcolor{green!50}2.02\\
\hline
\multirow{3}{*}{GSM8K} & $Exact-M \uparrow$ & 0.80 &\cellcolor{blue!35} 0.87 &\cellcolor{blue!35} 0.86 & 0.82 & 0.81 & 0.74 & 0.81 &\cellcolor{blue!30} 0.84 & 0.80 & 0.83 &\cellcolor{blue!30} 0.84 & 0.83 \\
& $Wh\downarrow$ & 13.21 & 14.44 & \cellcolor{red!50} 18.42 & \cellcolor{red!50}17.55  &\cellcolor{red!60} 20.28  & 10.38 & 13.55 & 12.78 & 13.26 & 12.80 & 12.98 & 13.50\\
\cline{2-14} 
& $ER\uparrow$ & 0.060 & 0.060 & 0.046 & 0.046 & 0.039 &\cellcolor{green!50} 0.071 & 0.059 &\cellcolor{green!50} 0.065 & 0.060 &\cellcolor{green!50} 0.064 &\cellcolor{green!50} 0.064 & 0.061\\
\hline
\multirow{3}{*}{HumanEval} & $pass@1 \uparrow$ & 0.71 &\cellcolor{blue!35} 0.81 &\cellcolor{blue!35} 0.79 & 0.63 & 0.65 & 0.69 & 0.74 & 0.72 & 0.54 & 0.62 & 0.74 &\cellcolor{blue!35} 0.78\\
& $Wh\downarrow$ & 4.31 & 4.85 &\cellcolor{red!35} 5.96 &\cellcolor{red!50} 8.08 & 4.52 & 3.16 & 4.39 & 3.83 & 4.87 &\cellcolor{red!35} 5.02 & 4.20 & 4.03\\
\cline{2-14} 
& $ER\uparrow$ & 0.164 & 0.167 & 0.132 & 0.077 & 0.143 &\cellcolor{green!50} 0.218 & 0.168 &\cellcolor{green!50} 0.187 & 0.110 & 0.123 & 0.176 &\cellcolor{green!50} 0.193\\
\hline
\multirow{3}{*}{Wikitext} & $MAUVE \uparrow$ & 87.74 & 91.93 & 85.22 & 86.14 &\cellcolor{blue!30} 92.9 & 92.32 &\cellcolor{blue!35} 96.95 &\cellcolor{blue!35} 94.55 & 88.87 & 92.76 & 90.12 & 92.53\\
& $Wh\downarrow$ & 71.32 & 73.94 & \cellcolor{red!40}82.87 & \cellcolor{red!50}102.47 & \cellcolor{red!40}80.41 & 55.26 & 71.18 & 71.26 & 70.49 & 70.91 & 70.07 & 70.93\\
\cline{2-14} 
& $ER\uparrow$ & 1.23 & 1.24 & 1.02 & 0.84 & 1.15 &\cellcolor{green!50} 1.67 &\cellcolor{green!50} 1.36 &\cellcolor{green!50} 1.32 & 1.26 & 1.30 & 1.28 & 1.30\\
\noalign{\hrule height 1.2pt}
\end{tabular}
 }
\vskip -0.1in
\end{table*}

In the HumanEval, Beam Search, DBS, and Min-p tasks, we achieve the highest pass@1 accuracy among all decoding methods. However, Min-p consumes much less energy, resulting in a higher energy efficiency ratio. In Wikitext, we observe that temperature sampling and Top-p sampling achieve the highest MAUVE scores, both exceeding 94, and at the same time rank among the strategies with the highest efficiency ratios, along with AD.

By considering all tasks and calculating the average ranking of decoding strategies based on quality and energy consumption in Qwen2.5 7B-Instruct, AD, Top-p, and Greedy perform best in terms of energy consumption, while Beam Search, Top-p, and DBS perform best in terms of achieved quality. Although AD has the highest efficiency ratio in all cases, in tasks like GSM8K its generative quality is relatively low, which lowers its overall quality ranking. CS, DoLa, and DBS have the highest amount of energy consumption compared to other methods. In contrast, stochastic sampling techniques like Top-p deliver near‐competitive generative quality scores at a fraction of the energy, leading to high efficiency ratios among the tested strategies, especially compared to beam-based deterministic strategies.

To better demonstrate how the choice of decoding strategy affects generative quality and energy consumption, we also employed the Relative Standard Deviation expressed as \(\text{RSD} (\%) = \frac{\sigma}{\mu} \times 100\) mathematically. RSD quantifies the variability of the data relative to its mean value. We utilize this statistical metric to provide a normalized measure of variability in text generation quality and energy consumption using various decoding methods across different tasks. Intuitively, a lower RSD indicates greater stability and consistency, while a higher value reflects greater variability. We depicted the RSD of quality and energy consumption of Qwen2.5 7B-Instruct for each task in Figure~\ref{rsd-dataset-qwen7B} (based on the results from Table~\ref{tab:performance_energy_comparison_7B}). It is clear that the selection of decoding strategies significantly affects the variability of energy consumption. Although in some cases, such as GSM8k or WikiText, we do not observe much variation in output quality when using different decoding methods, we do see significant energy deviations in such cases. This highlights the importance of being more energy-conscious when choosing a decoding strategy for our text generation tasks. Choosing an inappropriate strategy may not greatly reduce output quality, but it can lead to a significant increase in energy consumption. 

\begin{table*}[t]
\footnotesize
\centering
\caption{Relative Standard Deviation (RSD) and Sharpe Ratio (SR) for each decoding strategy in Qwen2.5-7B-Instruct. RSD is calculated for text generation quality and average energy consumption across different hyperparameter values. SR takes into account the quality and energy consumption of all hyperparameter runs within a decoding method. The table uses color coding to highlight key metrics: blue indicates low RSD in quality, red signifies high RSD in energy consumption, and yellow represents a high SR. $\uparrow$ indicates higher is better.}
\label{tab:rsd_energy_performance_hyper_7B}

\resizebox{0.99\textwidth}{!}{ 
\setlength{\tabcolsep}{3pt} 
\renewcommand{\arraystretch}{1.3} 
\begin{tabular}{!{\vrule width 1.2pt}>{\centering\arraybackslash}p{2cm}|c!{\vrule width 1.2pt}c|c|c|c|c|c!{\vrule width 1.2pt}c|c|c|c|c|c!{\vrule width 1.2pt}}
\noalign{\hrule height 1.2pt}  
\multirow{2}{*}{\textbf{Task}} & \multirow{2}{*}{\textbf{Metric}} & \multicolumn{6}{c!{\vrule width 1.2pt}}{\textbf{Deterministic Methods}} & \multicolumn{6}{c!{\vrule width 1.2pt}}{\textbf{Stochastic Methods}} \\
\cline{3-14}
& & \textbf{Beam} & \textbf{DBS} & \textbf{DoLa} & \textbf{CS(\scalebox{0.7}{k=5})} & \textbf{CS(\scalebox{0.7}{k=25})} & \textbf{AD} & \textbf{$\tau$} & \textbf{Top-p} & \textbf{Top-k} & \textbf{Eps} & \textbf{Typical} & \textbf{Min-p} \\
\noalign{\hrule height 1.2pt}
\multirow{3}{*}{De→En} & \textit{BL (RSD\%)} &\cellcolor{blue!55} 0.31 &\cellcolor{blue!55} 0.46 & 25.96 & 9.18 & 38.36 &\cellcolor{blue!55} 0.49 & 18.50 & 8.96 & 3.07 & 2.99 & 5.81 & 9.40 \\
& \textit{Wh (RSD\%)} & 13.32 & 11.39 &\cellcolor{red!65} 20.08 & 8.38 &\cellcolor{red!65} 35.30 & 2.25 & 5.18 & 2.88 & 1.22 & 1.53 & 2.70 & 3.09 \\
& $SR\uparrow$ & \(7.4e^{-3}\) & \(8.1e^{-3}\) & \(8.8e^{-4}\) & \(e^{-2}\) & \(1.3e^{-3}\) &\cellcolor{yellow!85} \(6e^{-2}\) & \(1.8e^{-2}\) & \(3.5e^{-2}\) &\cellcolor{yellow!85} \(6.5e^{-2}\) & \(5.1e^{-2}\) & \(3.3e^{-2}\) & \(3.4e^{-2}\) \\
\noalign{\hrule height 1.2pt}
\multirow{3}{*}{En→De} & \textit{BL (RSD\%)} &\cellcolor{blue!55} 1.30 &\cellcolor{blue!55} 2.16 & 28.86 & 18.19 & 51.47 & 5.41 & 13.42 & 13.32 & 8.04 &\cellcolor{blue!55} 2.18 & 4.38 & 11.62 \\
& \textit{Wh (RSD\%)} & 11.09 & 7.02 &\cellcolor{red!65} 16.47 & 3.96 &\cellcolor{red!65} 19.39 & 4.37 & 1.49 & 1.07 & 0.60 & 0.92 & 0.66 & 2.26 \\
& $SR\uparrow$ & \(4.5e^{-3}\) & \(6.8e^{-3}\) & \(9.4e^{-4}\) & \(e^{-2}\) & \(1.2e^{-3}\) & \(1.3e^{-2}\) & \(3.3e^{-2}\) & \(4.3e^{-2}\) & \cellcolor{yellow!85}\(5.3e^{-2}\) & \(3.5e^{-2}\) & \cellcolor{yellow!85}\(6.5e^{-2}\) & \(2.1e^{-2}\) \\
\noalign{\hrule height 1.2pt}
\multirow{3}{*}{GSM8K} & \textit{EM (RSD\%)} & 3.14 &\cellcolor{blue!55} 1.77 & 3.14 & 5.61 & 26.67 & 23.15 & 3.63 & 3.99 &\cellcolor{blue!55} 2.82 &\cellcolor{blue!55} 2.85 & 4.12 & 3.87 \\
& \textit{Wh (RSD\%)} &\cellcolor{red!65} 30.50 &\cellcolor{red!65} 20.06 & 0.31 & 4.81 & 17.78 & 8.61 & 1.52 & 0.50 & 0.90 & 1.27 & 1.74 & 1.18 \\
& $SR\uparrow$ & \(3.5e^{-5}\) & \(4.4e^{-5}\) &\cellcolor{yellow!85} \(2.4e^{-3}\) & \(1.9e^{-4}\) & \(1.7e^{-5}\) & \(1.9e^{-4}\) & \(e^{-3}\) &\cellcolor{yellow!85} \(2.3e^{-3}\) & \(1.5e^{-3}\) & \(1.1e^{-3}\) & \(8.8e^{-4}\) & \(e^{-3}\) \\
\noalign{\hrule height 1.2pt}
\multirow{3}{*}{HumanEval} & \textit{ACC (RSD\%)} &\cellcolor{blue!55} 1.79 &\cellcolor{blue!55} 1.40 & 11.50 & 38.86 & 68.65 & 9.96 & 23.94 & 9.73 &\cellcolor{blue!55} 2.12 & 7.00 & 11.56 & 12.10 \\
& \textit{Wh (RSD\%)} & \cellcolor{red!65}25.18 & 16.30 &\cellcolor{red!65} 45.95 & 11.02 &\cellcolor{red!65} 22.35 & 9.42 & 19.08 & 11.59 & 6.50 & 11.60 & 2.58 & 10.65 \\
& $SR\uparrow$ & \(1.2e^{-4}\) & \(1.6e^{-4}\) & \(2.3e^{-5}\) & \(2.6e^{-4}\) & \(5.5e^{-5}\) &\cellcolor{yellow!85} \(6.2e^{-4}\) & \(1.9e^{-4}\) & \(3.4e^{-4}\) & \(4.4e^{-4}\) & \(2.3e^{-4}\) &\cellcolor{yellow!85} \(1.5e^{-3}\) & \(4e^{-4}\) \\
\noalign{\hrule height 1.2pt}
\multirow{3}{*}{Wikitext} & \textit{MAU (RSD\%)} & 6.56 & 6.78 & 11.13 & 29.00 & 79.59 &\cellcolor{blue!55} 3.71 & 19.97 & 5.52 & 7.03 &\cellcolor{blue!55} 4.45 & \cellcolor{blue!55}4.23 & 6.91 \\
& \textit{Wh (RSD\%)} &\cellcolor{red!65} 8.01 &\cellcolor{red!65} 7.17 & 0.89 & 1.26 &\cellcolor{red!65} 5.15 & 2.05 & 0.90 & 0.87 & 0.60 & 1.05 & 0.80 & 1.25 \\
& $SR\uparrow$ & \(4.3e^{-3}\) & \(3.3e^{-3}\) & \(2.8e^{-2}\) & \(2e^{-2}\) & \(2e^{-3}\) & \(2.6e^{-2}\) & \(2.8e^{-2}\) & \(4.1e^{-2}\) & \cellcolor{yellow!85}\(6.3e^{-2}\) & \(3.3e^{-2}\) &\cellcolor{yellow!85} \(5.3e^{-2}\) & \(3.1e^{-2}\) \\
\noalign{\hrule height 1.2pt}
\end{tabular}  
} 
\vskip -0.1in
\end{table*}

\subsubsection{Overall Analysis of Decoding Strategies} 
\label{text_vs_energy_keyFindings_overal}

Based on the average ranking of decoding strategies across all tasks and models, AD consistently has the lowest energy consumption, followed by Greedy, Min-p, Top-p, and Temperature sampling. Beam Search, CS, DBS, and DoLa rank as the worst methods in terms of energy consumption. For the average ranking of generative quality, Beam Search, DBS, Top-p, and Min-p are the best-performing methods, while Epsilon, DoLA, and Top-k are the worst-performing ones. Although beam-based deterministic methods offer strong generative quality, they consume disproportionately high amounts of energy, calling into question their unnecessary use. Within a reasonable margin, stochastic decoding methods, when tuned to their optimal hyperparameters, show similar power consumption across tasks. This consistent pattern suggests that, despite using different sampling approaches, they impose comparable computational overhead during output generation, making them viable alternatives for each other in cases where only energy efficiency matters. Contrastive Search performs well in some tasks, but its high energy usage lowers its overall efficiency ratio. DoLa consistently ranks among the worst in terms of efficiency ratio. Although this method aims to reduce hallucinations and improve truthfulness in LLM outputs, it does not provide strong generative quality in our tasks and remains one of the most energy-consuming techniques.

\subsubsection{Per-Model Analysis of Decoding Strategies} 
\label{text_vs_energy_keyFindings_model}

Comparing the results of Qwen2.5-7B-Instruct with Llama3.1-8B-Instruct (Table \ref{tab:performance_energy_comparison_llama} in Appendix), we see that in both translation tasks, the energy consumed by the two models across most decoding strategies is relatively close. However, since Llama3.1 generally achieves higher generative quality, its efficiency ratios are larger in most cases. Interestingly, in GSM8K, Qwen2.5-7B shows much better generation accuracy in many cases, but still has lower efficiency ratios because it consumes more energy on this task compared to Llama3.1. If accuracy is prioritized, Qwen2.5-7B is the better choice, but if energy consumption is the focus, Llama3.1 is superior. In coding tasks, Qwen2.5-7B performs better in both quality and energy use for most decoding methods. For Wikitext, we see a similar trend: Qwen2.5-7B achieves slightly higher MAUVE scores while also using less energy, resulting in better efficiency ratios.

Comparing the results of Qwen2.5-7B-Instruct with Qwen2.5-14B-Instruct (Table \ref{tab:performance_energy_comparison_14B} in Appendix), we observe that although the 14B version achieves better generative performance in most tasks, its much higher energy consumption across all decoding methods leads to lower efficiency ratios compared to the 7B variant. On the Wikitext dataset, however, the 7B model delivers much higher MAUVE scores despite the 14B model consuming considerably more energy. This highlights that using larger models does not always guarantee better generative quality across all text generation scenarios.

Furthermore, looking at Figure \ref{rsd-dataset-all} and comparing the RSD of GPU energy consumption across different decoding strategies in the three tested models, we observe that the deviation in energy can be large in specific tasks like GSM8K and HumanEval for Llama3.1-8B and Qwen2.5-14B compared to Qwen2.5-7B. This shows that models from different architectures or sizes can behave quite differently in certain text generation tasks.

Regarding the ranking of decoding strategies based on quality and energy consumption in each model, Beam Search is consistently the best-performing strategy for generative quality across all three tested models. Top-p and DBS also rank among the best after Beam Search, while DoLA, epsilon, and Top-k sampling are always the weakest methods. For energy consumption, AD consistently achieves the best ranking, with Greedy and Min-p also performing well in terms of energy efficiency. In contrast, CS, DBS, and DoLA rank the worst for energy usage.

\subsubsection{Per-Task Analysis of Decoding Strategies} 
\label{text_vs_energy_keyFindings_task}

When evaluating the ranking of decoding strategies in each task, Beam Search and DBS consistently achieve the highest quality in both translation tasks and the HumanEval benchmark. In math problem-solving, Top-p and Beam Search perform best. DoLA, Epsilon, and Top-k rank the lowest in these tasks. In Wikitext, however, Top-k and Temperature sampling achieve the highest average ranks, while AD and DBS show the lowest rankings.

Regarding energy usage, AD consistently has the highest ranking across all tasks, while Greedy, Top-p, and Min-p are also among the best-performing strategies. In contrast, DBS, CS, DoLA, and Beam Search rank the lowest in terms of energy consumption.

Lastly, we observe that in specific tasks, such as German-to-English translation in Qwen2.5-14B or Llama3.1-8B, stochastic sampling strategies can sometimes consume more energy than Beam Search. This shows that energy usage is highly task- and model-dependent, and even sampling-based methods may use more energy than beam-based methods in certain cases.

\subsection{Hyperparameter Sensitivity Analysis} 

To investigate the impact of hyperparameter choices on both text generation quality and energy consumption, it is crucial to determine their sensitivity across different configurations. To quantify this, we calculate the RSDs of both quality and energy to measure their variability and consistency across all hyperparameters for each decoding strategy. In this way, we can capture the degree of sensitivity of each method to hyperparameter changes, allowing for an interpretable comparison between decoding methods in each task. In this scenario, a lower RSD suggests robustness, where hyperparameter changes have minimal effect on the observed metrics.

We also employed the Sharpe Ratio \cite{sharpe}, a classic statistical metric from finance, to incorporate the recorded energy variations observed in each run for a given hyperparameter within our analysis. Mathematically expressed as \(\frac{\mu - r_{f}}{\sigma}\), Sharpe Ratio measeures how well the average return of an investment compensates the volitality (risks) of returns. This reward-to-variability metric can be adapted in our analysis by treating the average of text generation qualities as the mean return of investments and the variability (standard deviation) of energy consumptions across all hyperparameter runs as the risks for each decoding strategy. This approach provides us a single metric to identify decoding strategies that are less sensitive to hyperparameter changes while simultaneously rewarding strong generative quality and penalizing energy fluctuations. A higher SR ratio signifies that the in all of our experiment runs for a specific decoding method, it consistently delivers good quality while minimizing energy consumption variations. Since we are ranking all decoding strategies within each task, we consider the risk‐free rate to be zero. Table \ref{tab:rsd_energy_performance_hyper_7B} presents the RSD of text generation quality and average energy consumption across all hyperparameter values for each decoding strategy, along with the SR between quality mean and energy consumption variations across all hyperparameter runs in Qwen2.5-7B-Instruct. The results for Llama3.1-8B-Instruct and Qwen2.5-14B-Instruct models are listed in Tables \ref{tab:rsd_energy_performance_hyper_8B} \& \ref{tab:rsd_energy_performance_hyper_14B} in Appendix respectively. 

Interestingly, Beam Search and DBS show relatively low RSD in quality for most tasks, especially in Translation and HumanEval across all the three tested models. However, they have relatively high RSD in energy, indicating stable generative performance but large variation in energy consumption when hyperparameters change. Although DoLA shows large variations in both quality and energy across its hyperparameters in most text generation tasks for the Qwen2.5 models, we observe that in the Llama3.1 model it remains impressively stable, with negligible change in either quality or energy. This highlights the point that the effect of hyperparameters on decoding strategies can differ greatly depending on the model family. Contrastive Search shows the highest variation in generation quality and energy usage across hyperparameters in all three models, especially when k is set to a large value, highlighting its strong sensitivity to hyperparameter tuning. Stochastic methods, except for Temperature sampling, are consistently among the strategies with the least variation in energy across different tasks and models. However, they show moderate fluctuations in generative quality across their hyperparameters. For Temperature sampling, we observe significantly high RSD values for both quality and energy in the Llama3.1 model. The reason Temperature scaling exhibits significant variation in both energy consumption and quality is the inclusion of high-temperature values in our hyperparameter choices. Higher temperatures \(\tau > 1\) increase randomness in token selection, reducing output fluency and coherence. Thus, the model struggles to enhance diversity without sacrificing accuracy, leading to a noticeable drop in quality and increased energy consumption. Surprisingly, in the Qwen2.5 variants, only quality shows large variation with temperature scaling, while energy remains relatively stable across the hyperparameter range. This highlights that large variations in generative quality do not always lead to significant fluctuations in energy consumption. Although in previous section we observed AD as the strategy with the lowest energy consumption in all tested scenarios, we find that it can show large variations in both quality and energy for tasks like GSM8K and HumanEval. In the other tasks we tested, both quality and energy remain stable across different hyperparameters for this deterministic strategy. 

\subsubsection{Key Observations}

Across all tasks and models, we observe that, except for Temperature sampling, the other stochastic decoding methods achieve some of the highest SR ratios, maintaining good quality with minimal energy fluctuations across different hyperparameter choices. This suggests that these methods are less sensitive to hyperparameter tuning and are more energy-conscious relative to the quality they produce.

In most cases, especially in translation and coding tasks, beam-based strategies (BS and DBS) exhibit significant energy fluctuations despite minimal variation in quality. Conversely, in translation tasks in Qwen2.5 variants, Temperature sampling exhibits the opposite trend, with substantial quality variation but stable energy fluctuations. This shows that variations in quality and energy across hyperparameters do not always correlate.

Another notable finding is that, while we identified AD as the most energy-saving method in our experiments, we observe that only in translation and coding tasks it ranks among the best methods with the highest SR ratio. In other cases, it exhibits variations in both generation quality and energy consumption across its selected hyperparameters, implying that AD performs optimally only when properly tuned in text generation tasks.

Contrastive Search stands out as the most sensitive to hyperparameter changes among all methods, with \(k=25\) exhibiting significantly great energy variations across different penalty \(\alpha\) values. Larger \( k\) values can lead to substantial shifts in both performance and energy consumption.

Additionally, we observe a unique pattern in the DoLA method. While Table \ref{tab:performance_energy_comparison_7B} shows that it is among the methods with the highest energy consumption, it also achieves some of the highest SR ratios in the Llama3.1-8B model. This indicates that even an energy-hungry decoding strategy may have minimal energy variation across different hyperparameters while still delivering acceptable (though not outstanding) generation quality in certain tasks, leading to high SR ratios.

Lastly, looking closely at the specific energy values for each hyperparameter, we observe that changing hyperparameters incrementally does not necessarily lead to a constant increase or decrease in energy. For example, when examining the average energy consumption of the min-p strategy across different hyperparameters in the HumanEval dataset with Qwen2.5-7B, we found that as the hyperparameter increases from 0.02 to 1.0, the energy fluctuates between 3.7 and 4.6 watt-hours, peaking at 5.4 watt-hours when the value is 0.05. This highlights that energy consumption does not always follow a smooth trend with hyperparameter changes, and small adjustments can sometimes lead to unexpected peaks or drops.

\section{Energy–Latency Relationship}

Since energy is calculated as the product of power and time, it depends not only on decoding latency but also on GPU power draw during measurement. When comparing two strategies, one with slightly higher runtime but lower power usage can still consume less energy overall, and vice versa. To check whether there is a difference between energy usage and wall-clock latency, we compared the energy ranking and latency ranking of decoding strategies across our tasks and models. We report the average decoding latency, average GPU power draw, and average GPU utilization for each decoding strategy in the Qwen2.5-7B-Instruct model in Table \ref{tab:latency_7B}. The corresponding results for the Llama3.1-8B-Instruct and Qwen2.5-14B-Instruct models are reported in Tables \ref{tab:latency_8B} and \ref{tab:latency_14B} in Appendix.

As expected, models of different sizes or architectures draw different ranges of power for each decoding strategy. For example, in Qwen2.5-7B the average power draw for Greedy ranges between 109–111 watts across all tasks, while in Llama3.1-8B it ranges between 105–111 watts, and in Qwen2.5-14B it ranges between 120–125 watts. When looking at each task separately, we also observe different power draws for different decoding strategies, despite running the experiments at the same application clock frequency. This is related to differences in GPU utilization and the varying computational complexity of each strategy in each task. In general, stochastic methods show stable and high GPU utilization, leading to relatively similar power draw across methods. In contrast, deterministic strategies show large deviations in GPU utilization and power usage. For instance, in Qwen2.5-7B in translation task, Beam Search has 91\% GPU utilization with a power draw of 102 watts, while DoLa reaches 96\% utilization with 125 watts, and CS has 88\% utilization with 103 watts of power draw.

\begin{figure}[h]
\centering
\includegraphics[width=0.7\textwidth]{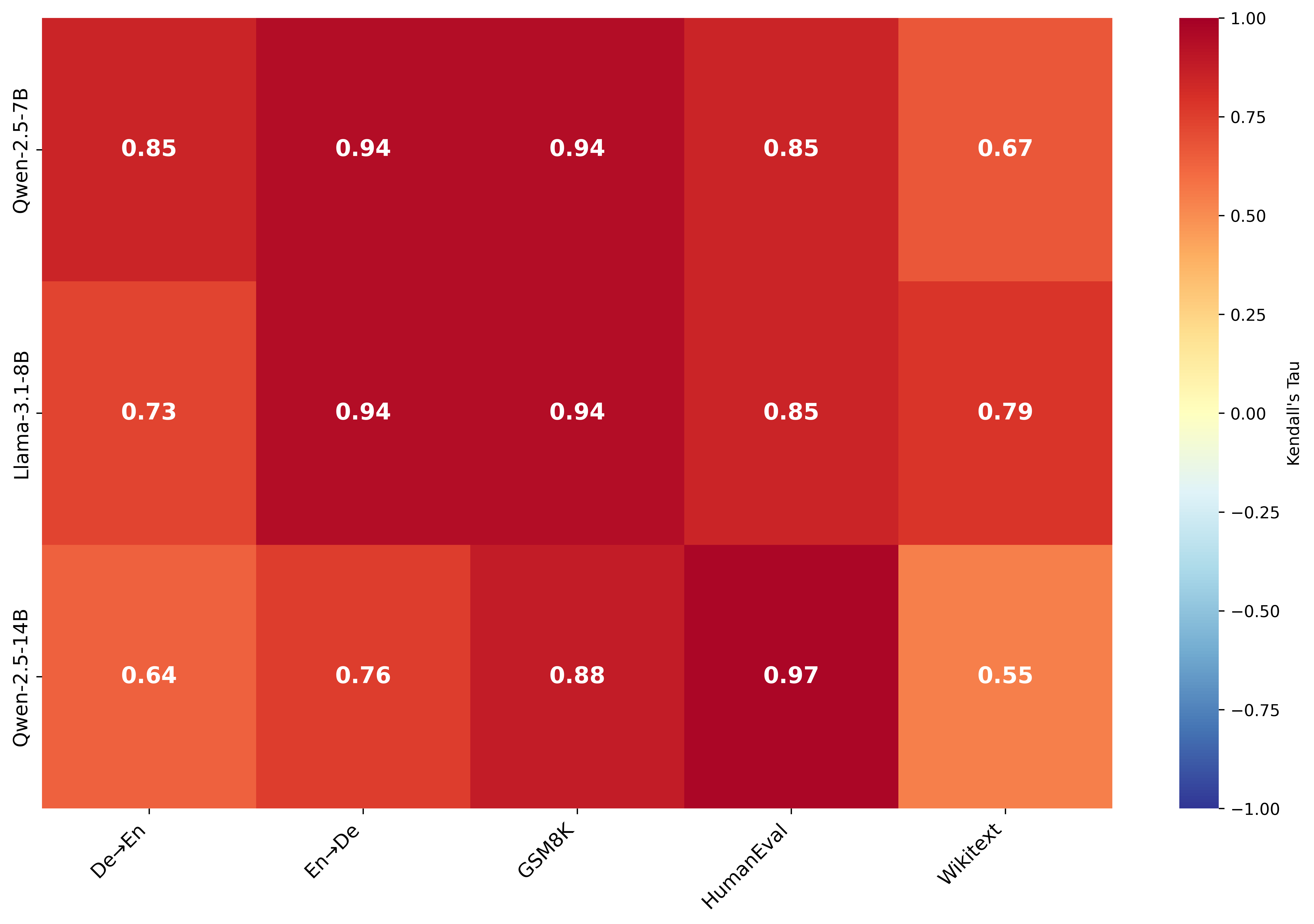}
\caption{Heatmap of Kendall’s Tau correlations between latency and energy consumption rankings of decoding strategies across different models and tasks.}
\label{fig:kendall}
\end{figure}

To measure the relationship between energy rankings and latency rankings of different decoding strategies across various models and tasks, we use Kendall's Tau correlation test. This non-parametric test is suitable because we have a relatively small sample size, the rankings do not follow a normal distribution, and the values are ordinal. The test shows how consistently energy and latency rankings are ordered in the same way across different cases. To better illustrate these results, we present them both as a heatmap (Fig \ref{fig:kendall}), which highlights the correlation patterns, and as a table of p-values (Table \ref{tab:kendall} in Appendix). 

Although in some cases, such as GSM8K and HumanEval with the Qwen2.5-7B and Llama3.1-8B models, the rankings are significantly correlated with minimal differences, in other cases—especially with Qwen2.5-14B—we observe smaller scores. For instance, in the German-to-English translation task, the test result is 0.64, compared to significantly higher values in other models. To better interpret these results, we analyze the energy and latency rankings. Interestingly, we find that Beam Search is one of the strategies with the largest deviation between its energy ranking and latency ranking. This deviation comes from its lower power draw of around 110 watts, compared to stochastic strategies that fluctuate around 120 watts. In this case, although Beam Search has higher latency, its lower power draw leads to lower overall energy consumption compared to strategies with lower latency but higher power draw. We observe similar situations in other cases where the energy rankings differ from the latency rankings of decoding strategies due to fluctuations in power draw. These observations show that energy consumption does not always align with latency due to differences in power draw among strategies. Understanding both is important for evaluating the overall efficiency of decoding methods.

\vspace{5mm}
\section{Limitations \& Future Work}
\label{limitation}

In this study, we confined our experiments to Hugging Face/PyTorch rather than inference frameworks like vLLM\cite{vllm} or SGLang\cite{sglang}. When conducting our study, we found that these frameworks did not support some of the decoding strategies we wanted to evaluate. Thus, to focus on a larger subset of decoding strategies, we decided to leave testing these optimized inference frameworks for future studies. While we carefully tried to select a large range of hyperparameters for our experiments, our approach was not exhaustive, leaving a broader range of parameter settings yet to be explored. We also did not combine different decoding strategies, suggesting another avenue for future research. Finally, our analysis focused on GPU energy consumption; however, future work could extend this scope to encompass other factors such as CPU power usage, memory and temperature, providing a more comprehensive understanding of how decoding strategies influence overall resource consumption. 

\section{Conclusion}
\label{conclusion}

In this study, we investigated the impact of decoding strategies on both generative quality and GPU energy consumption during LLM inference across various text generation tasks. Through a comprehensive evaluation, we found that the choice of decoding strategy significantly affects GPU energy consumption during inference, even in cases where it has minimal impact on output quality. Additionally, we observed that different decoding methods present distinct trade-offs between text quality and energy efficiency. While deterministic methods such as Beam Search and DBS generally achieve top-tier quality in most generation tasks, they also tend to consume more energy and exhibit higher energy fluctuations across their hyperparameters. In contrast, stochastic methods appear to be more energy-efficient, often providing a better quality-to-energy ratio and demonstrating lower sensitivity to hyperparameter changes relative to their output quality. Overall, no single decoding strategy outperforms all others across every metric. Instead, the optimal choice depends on the specific task and whether the priority is maximizing generation quality, minimizing energy consumption, or achieving a stable balance between both.

From a practical standpoint, our findings suggest the following general recommendations: Assisted Decoding (AD) consistently achieves the lowest energy consumption across all tasks, making it suitable for energy-constrained applications. Stochastic sampling strategies such as Top-p and Min-p offer a robust trade-off between quality and energy efficiency, providing stable and balanced performance across different workloads. Conversely, beam-based deterministic methods like Beam Search and DBS deliver the highest overall quality but at the cost of significantly greater energy usage. Therefore, AI practitioners aiming for energy-efficient inference should prioritize AD or stochastic sampling, while those emphasizing generative accuracy may prefer beam-based approaches.
\section{Ethical Considerations and Societal Impact} 

\begin{itemize}\itemsep0em 
    \item \textbf{Ethical Considerations:} This research does not involve any private or sensitive data, as all experiments are conducted using publicly available datasets. To ensure transparency and reproducibility, we rely on the LM-Evaluation-Harness framework, which provides a unified approach for benchmarking language models. However, the investigation of different text generation strategies must be accompanied by a careful assessment of potential ethical risks, including bias, misinformation, and harmful content generation.
    
    \item \textbf{Societal Impact:} This work aims to advance the field of Machine Learning while addressing the growing concerns about the energy consumption and environmental impact of large-scale AI models. Recently, there has been growing concern about the high energy consumption and sustainability challenges of AI models, emphasizing the need for efficient inference methods. Selecting decoding strategies that preserve model quality while reducing energy usage can help mitigate the environmental footprint of large-scale language models. Our research encourages the AI community to not only prioritize output quality but also adopt a more energy-conscious approach when using LLMs, contributing to greater sustainability in AI.
\end{itemize}
\section*{Code Availability}

All the code used in this study is publicly available in our GitHub repository \url{https://github.com/ds-anik/Energy-Conscious-LLM-Decoding}.

\section*{Acknowledgments}

The research presented in this paper has benefited from the Experimental Infrastructure for Exploration of Exascale Computing (eX3), which is financially supported by the Research Council of Norway under contract 270053.

\newpage

\appendix
\section{Appendix}
\label{appendix}

\vskip 0.5in
\begin{table*}[h]
\footnotesize
\centering
\caption{Optimal Hyperparameters for the Decoding Strategies used in the experiments with Qwen2.5-7B-Instruct. The Greedy decoding method does not require any hyperparameters, so we represented it with a dash (-) in the table.}
\label{tab:optimal_hyperparameter_7B}
\vskip 0.15in
\centering
\renewcommand{\arraystretch}{1.3} 
\setlength{\tabcolsep}{4pt} 
\begin{tabular}{!{\vrule width 1.2pt}c!{\vrule width 1.2pt}c|c|c|c|c|c!{\vrule width 1.2pt}c|c|c|c|c|c!{\vrule width 1.2pt}}
\noalign{\hrule height 1.2pt}
\multirow{2}{*}{\textbf{Dataset}} & \multicolumn{6}{c!{\vrule width 1.2pt}}{\textbf{Deterministic Methods}} & \multicolumn{6}{c!{\vrule width 1.2pt}}{\textbf{Stochastic Methods}} \\
\cline{2-13}
& \textbf{Greedy} & \textbf{Beam} & \textbf{DBS} & \textbf{DoLa} & \textbf{CS} & \textbf{AD} & $\tau$ & \textbf{Top-p} & \textbf{Top-k} & \textbf{Eps} & \textbf{Typical} & \textbf{Min-p} \\
\noalign{\hrule height 1.2pt}
De→En & - & 5 & (10,2) & "low" & \textnormal{k=25, $\alpha$=0.2} & 5 & 0.1 & 0.3 & 5 & 0.004 & 0.5 & 0.9\\
\hline
En→De & - & 10 & (10,2) & "low" & \textnormal{k=5, $\alpha$=0.2} & 2 & 0.1 & 0.3 & 5 & 0.004 & 0.5 & 1.0\\
\hline
GSM8K & - & 2 & (4,2) & "low" & \textnormal{k=5, $\alpha$=0.4} & 2 & 0.1 & 0.95 & 5 & 0.004 & 0.95 & 0.05 \\
\hline
HumanEval & - & 2 & (4,2) & "low" & \textnormal{k=5, $\alpha$=0.2} & 5 & 0.5 & 0.3 & 100 & 0.002 & 0.5 & 0.5\\
\hline
Wikitext & - & 2 & (4,2) & "low" & \textnormal{k=5, $\alpha$=0.2} & 5 & 0.3 & 0.7 & 50 & 0.0006 & 0.5 & 0.3\\
\noalign{\hrule height 1.2pt}
\end{tabular}
\vskip -0.1in
\end{table*}

\vskip 0.3in

\begin{table*}[h]
\footnotesize
\centering
\caption{Optimal Hyperparameters for the Decoding Strategies used in the experiments with Llama3.1-8B-Instruct. The Greedy decoding method does not require any hyperparameters, so we represented it with a dash (-) in the table.}
\label{tab:optimal_hyperparameter_8B}
\vskip 0.15in
\centering
\renewcommand{\arraystretch}{1.3} 
\setlength{\tabcolsep}{4pt} 
\begin{tabular}{!{\vrule width 1.2pt}c!{\vrule width 1.2pt}c|c|c|c|c|c!{\vrule width 1.2pt}c|c|c|c|c|c!{\vrule width 1.2pt}}
\noalign{\hrule height 1.2pt}
\multirow{2}{*}{\textbf{Dataset}} & \multicolumn{6}{c!{\vrule width 1.2pt}}{\textbf{Deterministic Methods}} & \multicolumn{6}{c!{\vrule width 1.2pt}}{\textbf{Stochastic Methods}} \\
\cline{2-13}
& \textbf{Greedy} & \textbf{Beam} & \textbf{DBS} & \textbf{DoLa} & \textbf{CS} & \textbf{AD} & $\tau$ & \textbf{Top-p} & \textbf{Top-k} & \textbf{Eps} & \textbf{Typical} & \textbf{Min-p} \\
\noalign{\hrule height 1.2pt}
De→En & - & 5 & (4,2) & "high" & \textnormal{k=5, $\alpha$=0.2} & 2 & 0.1 & 0.3 & 5 & 0.004 & 0.5 & 0.7\\
\hline
En→De & - & 10 & (4,2) & "low" & \textnormal{k=5, $\alpha$=0.2} & 2 & 0.1 & 0.3 & 5 & 0.004 & 0.5 & 0.9\\
\hline
GSM8K & - & 5 & (4,2) & "low" & \textnormal{k=5, $\alpha$=0.2} & 2 & 0.7 & 0.3 & 5 & 0.0009 & 0.8 & 0.7\\
\hline
HumanEval & - & 10 & (10,2) & "low" & \textnormal{k=5, $\alpha$=0.2} & 10 & 0.1 & 0.5 & 5 & 0.002 & 0.5 & 0.9\\
\hline
Wikitext & - & 2 & (4,2) & "low" & \textnormal{k=5, $\alpha$=0.4} & 2 & 0.7 & 0.5 & 50 & 0.002 & 0.9 & 0.02\\
\noalign{\hrule height 1.2pt}
\end{tabular}
\vskip -0.1in
\end{table*}

\vskip 0.3in

\begin{table*}[h]
\footnotesize
\centering
\caption{Optimal Hyperparameters for the Decoding Strategies used in the experiments with Qwen2.5-14B-Instruct. The Greedy decoding method does not require any hyperparameters, so we represented it with a dash (-) in the table.}
\label{tab:optimal_hyperparameter_14B}
\vskip 0.15in
\centering
\renewcommand{\arraystretch}{1.3} 
\setlength{\tabcolsep}{4pt} 
\begin{tabular}{!{\vrule width 1.2pt}c!{\vrule width 1.2pt}c|c|c|c|c|c!{\vrule width 1.2pt}c|c|c|c|c|c!{\vrule width 1.2pt}}
\noalign{\hrule height 1.2pt}
\multirow{2}{*}{\textbf{Dataset}} & \multicolumn{6}{c!{\vrule width 1.2pt}}{\textbf{Deterministic Methods}} & \multicolumn{6}{c!{\vrule width 1.2pt}}{\textbf{Stochastic Methods}} \\
\cline{2-13}
& \textbf{Greedy} & \textbf{Beam} & \textbf{DBS} & \textbf{DoLa} & \textbf{CS} & \textbf{AD} & $\tau$ & \textbf{Top-p} & \textbf{Top-k} & \textbf{Eps} & \textbf{Typical} & \textbf{Min-p} \\
\noalign{\hrule height 1.2pt}
De→En & - & 2 & (10,5) & "low" & \textnormal{k=5, $\alpha$=0.2} & 5 & 0.1 & 0.3 & 5 & 0.004 & 0.5 & 1.0\\
\hline
En→De & - & 10 & (10,2) & "low" & \textnormal{k=25, $\alpha$=0.2} & 2 & 0.1 & 0.3 & 5 & 0.004 & 0.5 & 0.9\\
\hline
GSM8K & - & 2 & (5,5) & "low" & \textnormal{k=25, $\alpha$=0.2} & 2 & 0.5 & 0.3 & 5 & 0.004 & 0.8 & 0.9\\
\hline
HumanEval & - & 10 & (10,2) & "low" & \textnormal{k=25, $\alpha$=0.2} & 10 & 0.5 & 0.7 & 5 & 0.004 & 0.8 & 0.5\\
\hline
Wikitext & - & 10 & (4,2) & "high" & \textnormal{k=5, $\alpha$=0.2} & 10 & 0.5 & 0.7 & 25 & 0.004 & 0.5 & 0.7 \\
\noalign{\hrule height 1.2pt}
\end{tabular}
\end{table*}

\begin{table*}[h]
\footnotesize
\centering
\caption{Text Generation Quality and Average Energy Consumption across Decoding Methods in their best Hyperparameter Setting using Llama3.1-8B-Instruct with batch size=1. The associated hyperparameters are listed in Table \ref{tab:optimal_hyperparameter_8B}. The table uses color coding to highlight key metrics: blue indicates high generative quality, red signifies high energy consumption, and green represents a high efficiency ratio. $\uparrow$ indicates higher is better, $\downarrow$ indicates lower is better}
\label{tab:performance_energy_comparison_llama}
\vskip 0.15in
\centering
\renewcommand{\arraystretch}{1.3} 
\resizebox{0.99\textwidth}{!}{
\setlength{\tabcolsep}{4pt} 
\begin{tabular}{!{\vrule width 1.2pt}c|c!{\vrule width 1.2pt}c|c|c|c|c|c!{\vrule width 1.2pt}c|c|c|c|c|c!{\vrule width 1.2pt}}
\noalign{\hrule height 1.2pt}
\multirow{2}{*}{\textbf{Task}} & \multirow{2}{*}{\textbf{Metric}} & \multicolumn{6}{c!{\vrule width 1.2pt}}{\textbf{Deterministic Methods}} & \multicolumn{6}{c!{\vrule width 1.2pt}}{\textbf{Stochastic Methods}} \\
\cline{3-14}
& & \textbf{Greedy} & \textbf{Beam} & \textbf{DBS} & \textbf{DoLa} & \textbf{CS} & \textbf{AD} & $\tau$ & \textbf{Top-p} & \textbf{Top-k} & \textbf{Eps} & \textbf{Typical} & \textbf{Min-p} \\
\noalign{\hrule height 1.2pt}
\multirow{3}{*}{De→En} & $BLEU \uparrow$ & 47.07 &\cellcolor{blue!35} 47.77 &\cellcolor{blue!35} 47.56 & 32.96 & 47.31 & 47.02 & 46.91 & \cellcolor{blue!35}47.50 & 31.94 & 33.49 & 38.31 & 47.18 \\
& $Wh\downarrow$ & 9.26 & 10.89 & 10.48 & \cellcolor{red!45}18.44 & 11.00 & 8.34 & 9.32 & 9.38 &\cellcolor{red!45} 11.02 & 10.50 &\cellcolor{red!45} 11.05 & 9.18 \\
\cline{2-14} 
& $ER\uparrow$ & \cellcolor{green!45}5.08 & 4.38 & 4.53 & 1.78 & 4.30 &\cellcolor{green!45} 5.63 & 5.03 &\cellcolor{green!45} 5.06 & 2.89 & 3.18 & 3.46 &\cellcolor{green!45} 5.13 \\
\hline
\multirow{3}{*}{En→De} & $BLEU \uparrow$ & 33.03 &\cellcolor{blue!35} 35.35 &\cellcolor{blue!35} 34.67 & 19.36 & 32.99 & 32.90 & 33.08 &\cellcolor{blue!35} 33.49 & 21.14 & 19.18 & 24.91 & 33.42 \\
& $Wh\downarrow$ & 13.40 &\cellcolor{red!45} 16.95 & 14.89 &\cellcolor{red!45} 27.10 &\cellcolor{red!45} 15.41 & 12.19 & 13.42 & 13.42 & 14.71 & 14.69 & 14.18 & 13.18 \\
\cline{2-14} 
& $ER\uparrow$ &\cellcolor{green!45} 2.46 & 2.08 & 2.32 & 0.71 & 2.14 &\cellcolor{green!45} 2.69 &\cellcolor{green!45} 2.46 &\cellcolor{green!45} 2.49 & 1.43 & 1.30 & 1.75 &\cellcolor{green!45} 2.53 \\
\hline
\multirow{3}{*}{GSM8K} & $Exact-M \uparrow$ & 0.77 &\cellcolor{blue!35} 0.78 & 0.76 & 0.66 &\cellcolor{blue!35} 0.78 & 0.75 & 0.75 &\cellcolor{blue!35} 0.80 & 0.61 & 0.62 & 0.66 & 0.76 \\
& $Wh\downarrow$ & 8.16 & \cellcolor{red!45}15.61 &\cellcolor{red!45} 13.85 &\cellcolor{red!45} 12.42 &\cellcolor{red!45} 13.58 & 6.54 & 8.15 & 8.36 & 8.39 & 9.57 & 8.73 & 7.86 \\
\cline{2-14} 
& $ER\uparrow$ &\cellcolor{green!45} 0.094 & 0.049 & 0.054 & 0.053 & 0.057 &\cellcolor{green!45} 0.114 & 0.092 &\cellcolor{green!45} 0.095 & 0.072 & 0.064 & 0.075 &\cellcolor{green!45} 0.096 \\
\hline
\multirow{3}{*}{HumanEval} & $pass@1 \uparrow$ & 0.71 &\cellcolor{blue!35} 0.78 &\cellcolor{blue!35} 0.77 & 0.67 & \cellcolor{blue!30}0.72 &\cellcolor{blue!30} 0.72 & 0.69 & 0.70 & 0.48 & 0.51 & 0.61 & 0.71\\
& $Wh\downarrow$ & 6.10 &\cellcolor{red!45} 16.89 &\cellcolor{red!45} 14.99 &\cellcolor{red!45} 8.48 & 7.33 & 5.27 & 6.13 & 4.85 & 7.67 & 8.13 & 6.98 & 6.03\\
\cline{2-14} 
& $ER\uparrow$ & 0.116 & 0.046 & 0.051 & 0.079 & 0.098 &\cellcolor{green!45} 0.136 & 0.112 &\cellcolor{green!45} 0.144 & 0.062 & 0.062 & 0.087 &\cellcolor{green!45} 0.117\\
\hline
\multirow{3}{*}{Wikitext} & $MAUVE \uparrow$ & 76.87 & 88.5 & 76.12 & 95.54 & 95.72 & 82.6 &\cellcolor{blue!35} 95.88 & 95.5 & \cellcolor{blue!35}97.37 & 95.73 &\cellcolor{blue!35} 96.95 & 95.2 \\
&$Wh\downarrow$ & 76.83 & \cellcolor{red!45}82.53 & \cellcolor{red!45}89.05 & \cellcolor{red!45}112.23 & \cellcolor{red!45}94.30 & 54.54 & 77.28 & 77.74 & 77.47 & 77.68 & 78.04 & 78.41\\
\cline{2-14} 
& $ER\uparrow$ & 1.00 & 1.07 & 0.85 & 0.85 & 1.01 &\cellcolor{green!45} 1.51 &\cellcolor{green!45} 1.24 & 1.22 &\cellcolor{green!45} 1.25 & 1.23 &\cellcolor{green!45} 1.24 & 1.21\\
\noalign{\hrule height 1.2pt}
\end{tabular}
 }
\vskip -0.1in
\end{table*}

\begin{table*}[h]
\footnotesize
\centering
\caption{Text Generation Quality and Average Energy Consumption across Decoding Methods in their best Hyperparameter Setting using Qwen2.5-14B-Instruct with batch size=1. The associated hyperparameters are listed in Table \ref{tab:optimal_hyperparameter_14B}. The table uses color coding to highlight key metrics: blue indicates high generative quality, red signifies high energy consumption, and green represents a high efficiency ratio. $\uparrow$ indicates higher is better, $\downarrow$ indicates lower is better }
\label{tab:performance_energy_comparison_14B}
\vskip 0.15in
\centering
\renewcommand{\arraystretch}{1.3} 
\resizebox{0.99\textwidth}{!}{
\setlength{\tabcolsep}{4pt} 
\begin{tabular}{!{\vrule width 1.2pt}c|c!{\vrule width 1.2pt}c|c|c|c|c|c!{\vrule width 1.2pt}c|c|c|c|c|c!{\vrule width 1.2pt}}
\noalign{\hrule height 1.2pt}
\multirow{2}{*}{\textbf{Task}} & \multirow{2}{*}{\textbf{Metric}} & \multicolumn{6}{c!{\vrule width 1.2pt}}{\textbf{Deterministic Methods}} & \multicolumn{6}{c!{\vrule width 1.2pt}}{\textbf{Stochastic Methods}} \\
\cline{3-14}
& & \textbf{Greedy} & \textbf{Beam} & \textbf{DBS} & \textbf{DoLa} & \textbf{CS} & \textbf{AD} & $\tau$ & \textbf{Top-p} & \textbf{Top-k} & \textbf{Eps} & \textbf{Typical} & \textbf{Min-p} \\
\noalign{\hrule height 1.2pt}
\multirow{3}{*}{De→En} & $BLEU \uparrow$ & 37.60 &\cellcolor{blue!35} 43.79 &\cellcolor{blue!35} 43.44 & 28.06 & 37.04 &\cellcolor{blue!35} 39.18 & 37.42 & 37.54 & 30.24 & 30.72 & 36.70 & 38.02 \\
& $Wh\downarrow$ & 19.41 & 18.44 &\cellcolor{red!45} 24.15 & \cellcolor{red!45}35.21 &\cellcolor{red!45} 22.80 & 16.44 & 19.47 & 19.55 & 19.98 & 20.06 & 18.85 & 19.11 \\
\cline{2-14} 
& $ER\uparrow$ & 1.93 &\cellcolor{green!45} 2.37 & 1.79 & 0.79 & 1.62 &\cellcolor{green!45} 2.38 & 1.92 & 1.92 & 1.51 & 1.53 &\cellcolor{green!45} 1.94 &\cellcolor{green!45} 1.98 \\
\hline
\multirow{3}{*}{En→De} & $BLEU \uparrow$ &\cellcolor{blue!25} 27.33 &\cellcolor{blue!35} 30.50 & \cellcolor{blue!35}30.32 & 23.78 & 27.20 & 27.24 &\cellcolor{blue!25} 27.32 &26.67 & 20.43 & 19.23 & 22.89 & 27.17 \\
& $Wh\downarrow$ & 23.80 &\cellcolor{red!35} 30.67 &\cellcolor{red!45} 31.10 &\cellcolor{red!45} 35.51 &\cellcolor{red!45} 35.04 & 22.47 & 24.24 & 23.86 & 24.34 & 24.81 & 23.88 & 23.93 \\
\cline{2-14} 
& $ER\uparrow$ &\cellcolor{green!45} 1.14 & 0.99 & 0.97 & 0.66 & 0.77 &\cellcolor{green!45} 1.21 &\cellcolor{green!45} 1.12 & 1.11 & 0.83 & 0.77 & 0.95 &\cellcolor{green!45} 1.13\\
\hline
\multirow{3}{*}{GSM8K} & $Exact-M \uparrow$ & 0.84 & 0.84 & 0.84 & 0.83 &\cellcolor{blue!35} 0.87 & 0.76 & 0.85 &\cellcolor{blue!35} 0.86 & 0.79 & 0.84 & 0.85 &\cellcolor{blue!35} 0.86 \\
& $Wh\downarrow$ & 37.07 &\cellcolor{red!45} 42.53 &\cellcolor{red!45} 69.54 &\cellcolor{red!45} 54.07 &\cellcolor{red!85} 192.8 & 25.14 & 38.91 & 37.58 & 36.49 & 36.62 & 37.70 & 37.32 \\
\cline{2-14} 
& $ER\uparrow$ & 0.022 & 0.019 & 0.012 & 0.015 & 0.004 &\cellcolor{green!45} 0.030 & 0.021 & 0.022 & 0.021 & 0.022 & 0.022 &\cellcolor{green!45} 0.023 \\
\hline
\multirow{3}{*}{HumanEval} & $pass@1 \uparrow$ & 0.69 &\cellcolor{blue!35} 0.87 &\cellcolor{blue!35} 0.86 & 0.66 & 0.69 & 0.68 & 0.73 &\cellcolor{blue!35} 0.78 & 0.58 & 0.65 & 0.72 & 0.71\\
& $Wh\downarrow$ & 6.04 &\cellcolor{red!45} 13.05 & \cellcolor{red!55}19.58 &\cellcolor{red!35} 10.67 &\cellcolor{red!45} 11.05 & 5.16 & 7.71 & 7.05 & 7.57 & 8.51 & 7.31 & 6.69\\
\cline{2-14} 
& $ER\uparrow$ &\cellcolor{green!45} 0.114 & 0.066 & 0.043 & 0.061 & 0.062 &\cellcolor{green!45} 0.131 & 0.094 &\cellcolor{green!45} 0.110 & 0.076 & 0.076 & 0.098 & 0.106\\
\hline
\multirow{3}{*}{Wikitext} & $MAUVE \uparrow$ & 74.93 & \cellcolor{blue!35}87.99 & 86.38 & 86.10 & 76.30 & 76.93 & 80.26 & 84.52 &\cellcolor{blue!35} 87.73 & 86.10 & 86.12 &\cellcolor{blue!35} 87.39\\
& $Wh\downarrow$ & 135.65 &\cellcolor{red!55} 186.74 &\cellcolor{red!55} 155.48 &\cellcolor{red!65} 199.43 &\cellcolor{red!55} 160.20 & 106.86 & 135.13 & 136.30 & 134.44 & 137.01 & 135.17 & 134.06\\
\cline{2-14} 
& $ER\uparrow$ & 0.55 & 0.47 & 0.55 & 0.43 & 0.47 &\cellcolor{green!45} 0.71 & 0.59 & 0.62 &\cellcolor{green!45} 0.65 & 0.62 & 0.63 &\cellcolor{green!45} 0.65\\
\noalign{\hrule height 1.2pt}
\end{tabular}
 }
\vskip -0.1in
\end{table*}

\begin{table*}[h]
\footnotesize
\centering
\caption{Relative Standard Deviation (RSD) and Sharpe Ratio (SR) for each decoding strategy in Llama3.1-8B-Instruct. RSD is calculated for text generation quality and average energy consumption across different hyperparameter values. SR takes into account the quality and energy consumption of all hyperparameter runs within a decoding method. $\uparrow$ indicates higher is better}
\label{tab:rsd_energy_performance_hyper_8B}
\vskip 0.15in
\resizebox{0.99\textwidth}{!}{ 
\setlength{\tabcolsep}{3pt} 
\renewcommand{\arraystretch}{1.3} 
\begin{tabular}{!{\vrule width 1.2pt}>{\centering\arraybackslash}p{2cm}|c!{\vrule width 1.2pt}c|c|c|c|c|c!{\vrule width 1.2pt}c|c|c|c|c|c!{\vrule width 1.2pt}}
\noalign{\hrule height 1.2pt}  
\multirow{2}{*}{\textbf{Task}} & \multirow{2}{*}{\textbf{Metric}} & \multicolumn{6}{c!{\vrule width 1.2pt}}{\textbf{Deterministic Methods}} & \multicolumn{6}{c!{\vrule width 1.2pt}}{\textbf{Stochastic Methods}} \\
\cline{3-14}
& & \textbf{Beam} & \textbf{DBS} & \textbf{DoLa} & \textbf{CS(\scalebox{0.7}{k=5})} & \textbf{CS(\scalebox{0.7}{k=25})} & \textbf{AD} & \textbf{$\tau$} & \textbf{Top-p} & \textbf{Top-k} & \textbf{Eps} & \textbf{Typical} & \textbf{Min-p} \\
\noalign{\hrule height 1.2pt}
\multirow{3}{*}{De→En} & \textit{BL (RSD\%)} & 0.34 & 0.51 & 2.29 & 12.80 & 34.92 & 1.67 & 34.97 & 13.71 & 4.18 & 4.07 & 7.31 & 12.53 \\
& \textit{Wh (RSD\%)} & 9.21 & 7.73 & 2.11 & 15.13 & 53.21 & 2.06 & 32.68 & 8.07 & 1.85 & 3.03 & 2.11 & 5.94 \\
& $SR\uparrow$ & \(1.2e^{-2}\) & \(1.4e^{-2}\) & \(2.2e^{-2}\) & \(6.3e^{-3}\) & \(8.3e^{-4}\) & \(7.4e^{-2}\) & \(2.5e^{-3}\) & \(1.3e^{-2}\) & \(3.8e^{-2}\) & \(2.5e^{-2}\) & \(4e^{-2}\) & \(1.9e^{-2}\) \\
\noalign{\hrule height 1.2pt}
\multirow{3}{*}{En→De} & \textit{BL (RSD\%)} & 0.69 & 1.53 & 0.15 & 7.06 & 22.00 & 4.71 & 45.50 & 18.17 & 8.08 & 4.87 & 14.24 & 16.80 \\
& \textit{Wh (RSD\%)} & 7.41 & 6.33 & 0.15 & 1.75 & 13.38 & 4.36 & 36.82 & 4.29 & 1.08 & 1.42 & 3.90 & 5.14 \\
& $SR\uparrow$ & \(8.5e^{-3}\) & \(8.9e^{-3}\) & \(7.8e^{-2}\) & \(2.9e^{-2}\) & \(2.6e^{-3}\) & \(1.6e^{-2}\) & \(e^{-3}\) & \(1.2e^{-2}\) & \(3e^{-2}\) & \(2.2e^{-2}\) & \(9.9e^{-3}\) & \(1.1e^{-2}\) \\
\noalign{\hrule height 1.2pt}
\multirow{3}{*}{GSM8K} & \textit{EM (RSD\%)} & 1.62 & 5.28 & 3.12 & 3.53 & 28.33 & 14.25 & 35.09 & 8.26 & 4.80 & 6.29 & 4.69 & 7.02 \\
& \textit{Wh (RSD\%)} & 33.02 & 24.71 & 2.66 & 15.00 & 40.85 & 8.68 & 32.70 & 1.37 & 6.02 & 0.37 & 1.84 & 2.92 \\
& $SR\uparrow$ & \(3.8e^{-5}\) & \(4.1e^{-5}\) & \(5.3e^{-4}\) & \(8.6e^{-5}\) & \(8e^{-6}\) & \(3.3e^{-4}\) & \(5.2e^{-5}\) & \(1.5e^{-3}\) & \(2.7e^{-4}\) & \(3e^{-3}\) & \(e^{-3}\) & \(7.8e^{-4}\) \\
\noalign{\hrule height 1.2pt}
\multirow{3}{*}{HumanEval} & \textit{ACC (RSD\%)} & 2.72 & 4.69 & 0.00 & 6.16 & 41.79 & 11.55 & 32.57 & 9.30 & 11.04 & 4.66 & 7.15 & 12.49 \\
& \textit{Wh (RSD\%)} & 50.41 & 43.17 & 0.63 & 10.38 & 80.34 & 18.34 & 94.56 & 17.19 & 9.05 & 8.95 & 11.12 & 15.66 \\
& $SR\uparrow$ & \(4.1e^{-5}\) & \(5.4e^{-5}\) & \(2.8e^{-3}\) & \(2.7e^{-4}\) & \(7e^{-6}\) & \(2.1e^{-4}\) & \(1.2e^{-5}\) & \(1.4e^{-4}\) & \(1.6e^{-4}\) & \(1.6e^{-4}\) & \(1.8e^{-4}\) & \(1.5e^{-4}\) \\
\noalign{\hrule height 1.2pt}
\multirow{3}{*}{Wikitext} & \textit{MAU (RSD\%)} & 6.92 & 7.56 & 0.60 & 3.49 & 35.98 & 2.04 & 44.56 & 1.12 & 2.32 & 2.09 & 2.37 & 8.59 \\
& \textit{Wh (RSD\%)} & 11.36 & 9.55 & 0.18 & 1.92 & 6.72 & 5.98 & 0.24 & 0.20 & 0.08 & 0.22 & 0.27 & 0.38 \\
& $SR\uparrow$ & \(2.3e^{-3}\) & \(2e^{-3}\) & \(15e^{-2}\) & \(2e^{-2}\) & \(2e^{-3}\) & \(7.8e^{-3}\) & \(9.8e^{-2}\) & \(10e^{-2}\) & \(37e^{-2}\) & \(12e^{-2}\) & \(9.4e^{-2}\) & \(7.2e^{-2}\) \\
\noalign{\hrule height 1.2pt}
\end{tabular}  
} 
\vskip -0.1in
\end{table*}

\begin{table*}[h]
\footnotesize
\centering
\caption{Relative Standard Deviation (RSD) and Sharpe Ratio (SR) for each decoding strategy in Qwen2.5-14B-Instruct. RSD is calculated for text generation quality and average energy consumption across different hyperparameter values. SR takes into account the quality and energy consumption of all hyperparameter runs within a decoding method. $\uparrow$ indicates higher is better}
\label{tab:rsd_energy_performance_hyper_14B}
\vskip 0.15in
\resizebox{0.99\textwidth}{!}{ 
\setlength{\tabcolsep}{3pt} 
\renewcommand{\arraystretch}{1.3} 
\begin{tabular}{!{\vrule width 1.2pt}>{\centering\arraybackslash}p{2cm}|c!{\vrule width 1.2pt}c|c|c|c|c|c!{\vrule width 1.2pt}c|c|c|c|c|c!{\vrule width 1.2pt}}
\noalign{\hrule height 1.2pt}  
\multirow{2}{*}{\textbf{Task}} & \multirow{2}{*}{\textbf{Metric}} & \multicolumn{6}{c!{\vrule width 1.2pt}}{\textbf{Deterministic Methods}} & \multicolumn{6}{c!{\vrule width 1.2pt}}{\textbf{Stochastic Methods}} \\
\cline{3-14}
& & \textbf{Beam} & \textbf{DBS} & \textbf{DoLa} & \textbf{CS(\scalebox{0.7}{k=5})} & \textbf{CS(\scalebox{0.7}{k=25})} & \textbf{AD} & \textbf{$\tau$} & \textbf{Top-p} & \textbf{Top-k} & \textbf{Eps} & \textbf{Typical} & \textbf{Min-p} \\
\noalign{\hrule height 1.2pt}
\multirow{3}{*}{De→En} & \textit{BL (RSD\%)} & 0.49 & 0.16 & 30.84 & 16.83 & 28.36 & 1.41 & 14.65 & 10.19 & 5.18 & 2.92 & 10.81 & 7.33 \\
& \textit{Wh (RSD\%)} & 10.30 & 7.61 & 26.00 & 18.41 & 34.15 & 1.84 & 4.40 & 4.81 & 2.02 & 2.33 & 5.99 & 2.22 \\
& $SR\uparrow$ & \(5.4e^{-3}\) & \(6.8e^{-3}\) & \(4.6e^{-4}\) & \(1.7e^{-3}\) & \(5.4e^{-4}\) & \(3.3e^{-2}\) & \(e^{-2}\) & \(9.6e^{-3}\) & \(1.7e^{-2}\) & \(1.6e^{-2}\) & \(7.2e^{-3}\) & \(2.2e^{-2}\) \\
\noalign{\hrule height 1.2pt}
\multirow{3}{*}{En→De} & \textit{BL (RSD\%)} & 1.48 & 2.62 & 20.19 & 14.05 & 45.42 & 4.69 & 24.81 & 12.85 & 7.60 & 3.68 & 8.44 & 10.49 \\
& \textit{Wh (RSD\%)} & 8.19 & 6.69 & 4.07 & 3.30 & 27.44 & 3.68 & 1.74 & 0.58 & 0.52 & 0.60 & 0.41 & 0.51 \\
& $SR\uparrow$ & \(3.5e^{-3}\) & \(4e^{-3}\) & \(3.5e^{-3}\) & \(6.8e^{-3}\) & \(4.2e^{-4}\) & \(8.8e^{-3}\) & \(1.3e^{-2}\) & \(4.2e^{-2}\) & \(3.5e^{-2}\) & \(3.2e^{-2}\) &\(4.8e^{-2}\)  & \(4.9e^{-2}\) \\
\noalign{\hrule height 1.2pt}
\multirow{3}{*}{GSM8K} & \textit{EM (RSD\%)} & 2.28 & 1.00 & 0.60 & 1.47 & 54.54 & 26.74 & 3.85 & 2.60 & 1.05 & 6.21 & 2.81 & 2.65 \\
& \textit{Wh (RSD\%)} & 35.11 & 23.10 & 0.61 & 0.88 & 1.75 & 9.36 & 0.62 & 1.15 & 0.16 & 0.50 & 0.99 & 1.09 \\
& $SR\uparrow$ & \(9e^{-6}\) & \(1.2e^{-5}\) & \(6.4e^{-4}\) & \(3.6e^{-4}\) & \(4.7e^{-5}\) & \(7.4e^{-5}\) & \(8e^{-4}\) & \(4.7e^{-4}\) & \(1.1e^{-3}\) & \(7.5e^{-4}\) & \(5.3e^{-4}\) & \(5.3e^{-4}\) \\
\noalign{\hrule height 1.2pt}
\multirow{3}{*}{HumanEval} & \textit{ACC (RSD\%)} & 2.43 & 3.56 & 12.82 & 9.81 & 44.21 & 4.73 & 25.00 & 8.11 & 4.37 & 9.31 & 10.18 & 8.23 \\
& \textit{Wh (RSD\%)} & 22.65 & 25.40 & 45.22 & 5.44 & 77.28 & 13.97 & 34.15 & 8.75 & 4.97 & 9.38 & 4.42 & 12.05 \\
& $SR\uparrow$ & \(e^{-4}\) & \(6.5e^{-5}\) & \(1.8e^{-5}\) & \(4.1e^{-4}\) & \(7e^{-6}\) & \(2.7e^{-4}\) & \(5.9e^{-5}\) & \(3.2e^{-4}\) & \(3.9e^{-4}\) & \(1.9e^{-4}\) & \(5.1e^{-4}\) & \(2.2e^{-4}\) \\
\noalign{\hrule height 1.2pt}
\multirow{3}{*}{Wikitext} & \textit{MAU (RSD\%)} & 5.15 & 5.40 & 1.41 & 21.53 & 65.05 & 2.24 & 4.44 & 5.04 & 3.84 & 0.89 & 5.39 & 4.61 \\
& \textit{Wh (RSD\%)} & 10.98 & 8.71 & 0.26 & 0.68 & 1.90 & 2.03 & 0.40 & 0.45 & 0.67 & 0.48 & 0.42 & 0.88 \\
& $SR\uparrow$ & \(1.3e^{-3}\) & \(1.5e^{-3}\) & \(5.5e^{-2}\) & \(1.4e^{-2}\) & \(2.3e^{-3}\) & \(1.2e^{-2}\) & \(4e^{-2}\) & \(4.5e^{-2}\) & \(3.2e^{-2}\) & \(4.1e^{-2}\) &\(4.2e^{-2}\)  & \(1.9e^{-2}\) \\
\noalign{\hrule height 1.2pt}
\end{tabular}  
} 
\vskip -0.1in
\end{table*}

\begin{table*}[h]
\footnotesize
\centering
\caption{Average Energy Consumption (watt-hour per query), Average Latency (seconds per query), Average GPU Power Draw (watts) and Average GPU Utilization (\%) in Qwen2.5-7B-Instruct model in the best Hyperparameter for each method}
\label{tab:latency_7B}
\vskip 0.15in
\resizebox{0.93\textwidth}{!}{ 
\setlength{\tabcolsep}{3pt} 
\renewcommand{\arraystretch}{1.3} 
\begin{tabular}{!{\vrule width 1.2pt}>{\centering\arraybackslash}p{2cm}|c!{\vrule width 1.2pt}c|c|c|c|c|c!{\vrule width 1.2pt}c|c|c|c|c|c!{\vrule width 1.2pt}}
\noalign{\hrule height 1.2pt}  
\multirow{2}{*}{\textbf{Task}} & \multirow{2}{*}{\textbf{Metric}} & \multicolumn{6}{c!{\vrule width 1.2pt}}{\textbf{Deterministic Methods}} & \multicolumn{6}{c!{\vrule width 1.2pt}}{\textbf{Stochastic Methods}} \\
\cline{3-14}
& & \textbf{Greedy} & \textbf{Beam} & \textbf{DBS} & \textbf{DoLa} & \textbf{CS} & \textbf{AD} & \textbf{$\tau$} & \textbf{Top-p} & \textbf{Top-k} & \textbf{Eps} & \textbf{Typical} & \textbf{Min-p} \\
\noalign{\hrule height 1.2pt}
\multirow{4}{*}{De→En} & \textit{Wh/query} &0.0187 &0.0222 &0.0284 &0.0445 &0.0265 &0.0164 & 0.0190 &0.0187 & 0.0205& 0.0201 & 0.0201 & 0.0189 \\
& \textit{sec/query} & 0.609& 0.784& 1.030& 1.28& 0.927& 0.555& 0.624& 0.627& 0.673& 0.677& 0.675& 0.619 \\
& \textit{ util (\%)} & 95.2 & 91.0 & 89.0 & 96.0 & 88.0 & 92.0 & 94.0 & 95.0 & 94.2 & 95.0 & 95.0 & 94.2 \\
& \textit{watt} & 111.07 & 102.11 & 99.49 & 125.15 & 103.02 & 106.73 & 110.02 & 107.56 & 110.07 & 107.28 & 107.19 & 110.03 \\
\noalign{\hrule height 1.2pt}
\multirow{4}{*}{En→De} & \textit{Wh/query} &0.0260 & 0.0357&0.0348 &0.0437 & 0.0308&0.0241 &0.0262 &0.0264 &0.0268 &0.0273 & 0.0270 & 0.0265\\
& \textit{sec/query} & 0.838 & 1.28 & 1.26& 1.25& 1.14& 0.821& 0.860& 0.876& 0.886& 0.920& 0.912& 0.878 \\
& \textit{ util (\%)} & 96.0 & 91.0 & 88.8 & 96.0 & 85.8 & 92.6 & 94.6 & 95.0 & 95.4 & 95.0 & 95.0 & 93.4 \\
& \textit{watt} & 111.72 & 100.37& 99.3& 125.22& 96.64& 105.8& 110.02& 108.78& 109.03& 106.93& 106.89& 108.9 \\
\noalign{\hrule height 1.2pt}
\multirow{4}{*}{GSM8K} & \textit{Wh/query} &0.132 &0.144 &0.184 &0.175 &0.202 &0.103 &0.135 &0.127 &0.132 &0.128 &0.129 &0.135 \\ 
& \textit{sec/query} & 4.31& 4.75& 5.94& 5.17& 6.79& 3.81& 4.31& 4.18 & 4.33& 4.19& 4.19& 4.42\\
& \textit{ util (\%)} & 95.4 & 93.2 & 93.0 & 96.2 & 90.0 & 92.2 & 96.0 & 96.0 & 96.0 & 96.0 & 96.0 & 95.6 \\
& \textit{watt} & 110.32& 109.39& 111.55& 122.25& 107.47& 97.97& 113.02& 110.0& 110.19& 109.85& 111.35& 109.88 \\
\noalign{\hrule height 1.2pt}
\multirow{4}{*}{HumanEval} & \textit{Wh/query} &0.0431 & 0.0484&0.0596 &0.0808 &0.0452 &0.0316 &0.0438 &0.0383 &0.0487 &0.0501 &0.042 &0.0403 \\
& \textit{sec/query} & 1.41& 1.72& 2.15& 2.41& 1.72& 1.11& 1.46& 1.29& 1.64& 1.68& 1.44& 1.35 \\
& \textit{ util (\%)} & 94.0 & 92.0 & 90.0 & 95.2 & 86.2 & 90.0 & 94.0 & 93.8 & 94.4 & 94.0 & 94.2 & 93.2 \\
& \textit{watt} & 109.71& 101.33& 99.81& 120.66& 94.44& 102.19& 108.27& 106.78& 106.78& 107.51& 104.61& 106.74 \\
\noalign{\hrule height 1.2pt}
\multirow{4}{*}{Wikitext} & \textit{Wh/query} &0.1783 &0.1848 &0.2071 &0.2561 &0.2010 &0.1381 &0.1779 &0.1781 &0.1762 &0.1772 &0.1751 &0.1773 \\
& \textit{sec/query} & 5.77& 6.28& 7.27& 7.55& 7.49& 4.74& 5.83& 5.92& 5.82& 5.81& 5.83& 5.90 \\
& \textit{ util (\%)} & 96.0 & 96.6 & 92.8 & 97.0 & 89.0 & 94.6 & 96.0 & 96.6 & 96.4 & 96.4 & 96.6 & 96.4 \\
& \textit{watt} & 111.18& 105.92& 102.55& 122.12& 96.51& 104.73& 109.78& 108.17& 108.89& 109.79& 108.14& 108.11 \\
\noalign{\hrule height 1.2pt}
\end{tabular}  
} 
\vskip -0.1in
\end{table*}

\begin{table*}[h]
\footnotesize
\centering
\caption{Average Energy Consumption (watt-hour per query), Average Latency (seconds per query), Average GPU Power Draw (watts) and Average GPU Utilization (\%) in Llama-3.1-8B-Instruct model in the best Hyperparameter for each method}
\label{tab:latency_8B}
\vskip 0.15in
\resizebox{0.93\textwidth}{!}{ 
\setlength{\tabcolsep}{3pt} 
\renewcommand{\arraystretch}{1.3} 
\begin{tabular}{!{\vrule width 1.2pt}>{\centering\arraybackslash}p{2cm}|c!{\vrule width 1.2pt}c|c|c|c|c|c!{\vrule width 1.2pt}c|c|c|c|c|c!{\vrule width 1.2pt}}
\noalign{\hrule height 1.2pt}  
\multirow{2}{*}{\textbf{Task}} & \multirow{2}{*}{\textbf{Metric}} & \multicolumn{6}{c!{\vrule width 1.2pt}}{\textbf{Deterministic Methods}} & \multicolumn{6}{c!{\vrule width 1.2pt}}{\textbf{Stochastic Methods}} \\
\cline{3-14}
& & \textbf{Greedy} & \textbf{Beam} & \textbf{DBS} & \textbf{DoLa} & \textbf{CS} & \textbf{AD} & \textbf{$\tau$} & \textbf{Top-p} & \textbf{Top-k} & \textbf{Eps} & \textbf{Typical} & \textbf{Min-p} \\
\noalign{\hrule height 1.2pt}
\multirow{4}{*}{De→En} & \textit{Wh/query} &0.0185 &0.0217 &0.0209 &0.0368 &0.022 &0.0166 &0.0186 &0.0187 &0.022 &0.021 &0.0221 &0.0183 \\
& \textit{sec/query} & 0.630 & 0.768& 0.776& 1.09 &0.835& 0.584 & 0.638& 0.644& 0.753& 0.719& 0.760& 0.632 \\
& \textit{ util (\%)} & 91.0 & 91.0 & 89.0 & 92.8 & 86.8 & 92.2 & 91.4 & 91.4 & 95.6 & 95.4 & 91.2 & 95.2 \\
& \textit{watt} & 105.73& 102.09& 97.2& 120.83& 94.82& 102.87& 105.04& 104.89& 105.33& 105.07& 104.6& 104.45 \\
\noalign{\hrule height 1.2pt}
\multirow{4}{*}{En→De} & \textit{Wh/query} &0.0268 &0.0339 &0.0297 &0.0542 &0.0308 &0.0243 &0.0268 &0.0268 &0.0294 &0.0293 &0.0283 & 0.0263\\
& \textit{sec/query} & 0.911& 1.17& 1.10& 1.61& 1.18& 0.854& 0.917& 0.921& 1.00& 1.00 & 0.978& 0.904 \\
& \textit{ util (\%)} & 91.0 & 91.4 & 89.0 & 92.8 & 87.0 & 92.4 & 91.2 & 91.6 & 95.2 & 95.8 & 95.0 & 95.4 \\
& \textit{watt} & 105.9& 103.63& 96.88& 120.66& 93.66& 102.83& 105.33& 104.89& 105.78& 105.25& 104.36& 105.0 \\
\noalign{\hrule height 1.2pt}
\multirow{4}{*}{GSM8K} & \textit{Wh/query} &0.0816 &0.1561 &0.1384 &0.1242 &0.1358 &0.0654 &0.0815 &0.0836 &0.0839 &0.0957 &0.0873 &0.0786 \\
& \textit{sec/query} & 2.63& 4.68& 4.44& 3.61& 4.47& 2.39& 2.71& 2.74& 2.81& 3.19& 2.945 & 2.63 \\
& \textit{ util (\%)} & 96.0 & 95.0 & 93.0 & 97.0 & 91.0 & 92.4 & 96.2 & 96.0 & 96.0 & 96.8 & 96.6 & 97.0 \\
& \textit{watt} & 111.44& 119.9& 112.27& 123.83& 109.35& 98.41& 108.16& 109.63& 107.23& 107.86& 106.81& 107.64\\
\noalign{\hrule height 1.2pt}
\multirow{4}{*}{HumanEval} & \textit{Wh/query} &0.0613 &0.1689 &0.1499 &0.0848&0.0733 &0.0527 &0.0613 &0.0484 &0.0767 &0.0813 &0.0698 &0.0603 \\
& \textit{sec/query} & 2.06& 5.79& 5.02& 2.60& 2.78& 1.93& 2.11& 1.67& 2.66& 2.79& 2.44& 2.08 \\
& \textit{ util (\%)} & 93.8 & 86.0 & 91.6 & 96.0 & 87.0 & 91.4 & 93.2 & 93.6 & 94.8 & 93.0 & 94.8 & 93.2 \\
& \textit{watt} & 106.27& 104.97& 107.48& 117.44& 94.79& 98.32& 104.51& 104.34& 103.78& 104.7& 102.72& 104.5 \\
\noalign{\hrule height 1.2pt}
\multirow{4}{*}{Wikitext} & \textit{Wh/query} &0.1920 &0.2063 &0.2226 & 0.2805&0.2357 &0.1363 &0.1932 &0.1943 & 0.1936&0.1942 &0.1951 &0.1960 \\
& \textit{sec/query} & 6.48 & 7.11& 8.02& 8.25& 8.88& 4.70& 6.45& 6.55 & 6.58& 6.52& 6.58& 6.53 \\
& \textit{ util (\%)} & 96.6 & 97.0 & 94.0 & 97.0 & 88.6 & 94.8 & 96.8 & 96.4 & 96.6 & 96.6 & 96.2 & 96.4 \\
& \textit{watt} & 106.66& 104.42& 99.89& 122.38& 95.57& 104.4& 107.78& 106.77& 105.98& 107.09& 106.64& 108.04\\
\noalign{\hrule height 1.2pt}
\end{tabular}  
} 
\vskip -0.1in
\end{table*}

\begin{table*}[h]
\footnotesize
\centering
\caption{Average Energy Consumption (watt-hour per query), Average Latency (seconds per query), Average GPU Power Draw (watts) and Average GPU Utilization (\%) in Qwen2.5-14B-Instruct model in the best Hyperparameter for each method}
\label{tab:latency_14B}

\resizebox{0.93\textwidth}{!}{ 
\setlength{\tabcolsep}{3pt} 
\renewcommand{\arraystretch}{1.3} 
\begin{tabular}{!{\vrule width 1.2pt}>{\centering\arraybackslash}p{2cm}|c!{\vrule width 1.2pt}c|c|c|c|c|c!{\vrule width 1.2pt}c|c|c|c|c|c!{\vrule width 1.2pt}}
\noalign{\hrule height 1.2pt}  
\multirow{2}{*}{\textbf{Task}} & \multirow{2}{*}{\textbf{Metric}} & \multicolumn{6}{c!{\vrule width 1.2pt}}{\textbf{Deterministic Methods}} & \multicolumn{6}{c!{\vrule width 1.2pt}}{\textbf{Stochastic Methods}} \\
\cline{3-14}
& & \textbf{Greedy} & \textbf{Beam} & \textbf{DBS} & \textbf{DoLa} & \textbf{CS} & \textbf{AD} & \textbf{$\tau$} & \textbf{Top-p} & \textbf{Top-k} & \textbf{Eps} & \textbf{Typical} & \textbf{Min-p} \\
\noalign{\hrule height 1.2pt}
\multirow{4}{*}{De→En} & \textit{Wh/query} &0.0388 &0.0368&0.0482 &0.0704 &0.0456 &0.0328 &0.0389 &0.0391 &0.0399 & 0.0401&0.0377 & 0.0382\\
& \textit{sec/query} & 1.12 & 1.19& 1.58& 1.87& 1.58& 0.99& 1.13& 1.15& 1.18& 1.18& 1.12& 1.14 \\
& \textit{util (\%)} & 97.0 & 93.0 & 90.0 & 98.0 & 88.0 & 94.8 & 96.4 & 96.6 & 97.4 & 96.4 & 97.4 & 97.6 \\
& \textit{watt} & 124.53& 110.86& 109.74& 134.93& 103.56& 119.58& 123.2& 122.44& 121.02& 122.25& 120.24& 120.48 \\
\noalign{\hrule height 1.2pt}
\multirow{4}{*}{En→De} & \textit{Wh/query} & 0.0476& 0.0613&0.0622 &0.0710 &0.0700 &0.0449 &0.0484 &0.0477 &0.0486 &0.0496 &0.0477 &0.0478 \\
& \textit{sec/query} & 1.40 & 1.96 & 1.99 & 1.88 & 2.29 & 1.34 & 1.42 & 1.42 & 1.45 & 1.46 & 1.42 & 1.42 \\
& \textit{util (\%)} & 97.4 & 94.0 & 92.0 & 98.0 & 91.0 & 94.8 & 96.0 & 97.0 & 97.2 & 96.8 & 97.6 & 97.4 \\
& \textit{watt} & 122.0& 112.66& 112.18& 135.39& 109.74& 120.21& 122.7& 120.37& 120.29& 122.37& 120.58& 120.82\\
\noalign{\hrule height 1.2pt}
\multirow{4}{*}{GSM8K} & \textit{Wh/query} &0.3707 &0.4253 &0.6954 &0.5407 &1.928 &0.2514 &0.3890 &0.3757 & 0.3649&0.3661 & 0.377&0.3732 \\
& \textit{sec/query} & 10.65& 13.07& 20.28& 14.52& 50.26 & 8.31 & 11.19 & 10.85 & 10.81 & 10.84 & 11.18 &  10.80\\
& \textit{util (\%)} & 97.2 & 97.0 & 94.4 & 98.0 & 97.0 & 95.0 & 97.2 & 97.6 & 97.4 & 98.0 & 98.0 & 97.6 \\
& \textit{watt} & 125.25& 117.07& 123.4& 134.02& 138.11& 108.85& 125.16& 124.66& 121.47& 121.63& 121.41& 124.42 \\
\noalign{\hrule height 1.2pt}
\multirow{4}{*}{HumanEval} & \textit{Wh/query} &0.0604 &0.1305 &0.1957 &0.1067 &0.1105 &0.0516 &0.0771 &0.0705 &0.0757 &0.0851 &0.0731 &0.0669 \\
& \textit{sec/query} & 1.80 & 4.06 & 6.01 & 2.83 & 3.48 & 1.66 & 2.29 & 2.16 & 2.31 & 2.55 & 2.19 & 2.04 \\
& \textit{ util (\%)} & 96.0 & 91.8 & 94.0 & 97.0 & 91.8 & 94.0 & 96.2 & 96.0 & 96.4 & 96.0 & 96.4 & 96.4 \\
& \textit{watt} & 120.69& 115.64& 117.12& 135.37& 114.3& 112.02& 120.83& 117.38& 117.81& 120.05& 119.91& 117.7 \\
\noalign{\hrule height 1.2pt}
\multirow{4}{*}{Wikitext} & \textit{Wh/query} & 0.3391& 0.4668&0.3887 &0.4985 &0.4004 &0.2671 &0.3378 &0.3407 &0.3361 &0.3425 &0.3379 &0.3351 \\
& \textit{sec/query} & 9.90 & 14.13 & 12.47 & 13.43 & 13.64 & 8.36 & 9.91 & 10.03 & 10.09 & 10.08 & 10.26 & 10.12 \\
& \textit{ util (\%)} & 97.6 & 98.4 & 96.0 & 98.0 & 89.8 & 96.6 & 97.6 & 98.0 & 98.0 & 98.0 & 97.2 & 97.6 \\
& \textit{watt} & 123.3& 118.95& 112.17& 133.6& 105.66& 115.01& 122.7& 122.24& 119.88& 122.29& 118.5& 119.17 \\
\noalign{\hrule height 1.2pt}
\end{tabular}  
} 
\vskip -0.1in
\end{table*}

\begin{table*}[h]
\footnotesize
\centering
\caption{Per-Model Friedman Test Statistics for Quality and Energy Consumption values across 5 tasks ($df=11$)}
\label{tab:friedman_model}

\centering
\renewcommand{\arraystretch}{1.3} 
\setlength{\tabcolsep}{4pt} 
\begin{tabular}{!{\vrule width 1.2pt}c!{\vrule width 1.2pt}c|c!{\vrule width 1.2pt}c|c!{\vrule width 1.2pt}}
\noalign{\hrule height 1.2pt}
\multirow{2}{*}{\textbf{Model}} & \multicolumn{2}{c!{\vrule width 1.2pt}}{\textbf{Quality}} & \multicolumn{2}{c!{\vrule width 1.2pt}}{\textbf{Energy}} \\
\cline{2-5}
& $X^2$ & $p$ & $X^2$ & $p$ \\
\noalign{\hrule height 1.2pt}
Qwen-2.5-7B-Inst & 28.37 & 0.00284 & 43.15 & $1.02e^{-05}$ \\
\hline
Llama-3.1-8B-Inst & 21.27 & 0.0307 & 43.96 & $7.38e^{-06}$ \\
\hline
Qwen-2.5-14B-Inst & 23.35 & 0.0158 & 41.80 & $1.75e^{-05}$ \\
\noalign{\hrule height 1.2pt}
\end{tabular}
\vskip -0.1in
\end{table*}

\begin{table*}[h]
\footnotesize
\centering
\caption{Per-Task Friedman Test Statistics for Quality and Energy Consumption values across 3 models ($df=11$)}
\label{tab:friedman_task}

\centering
\renewcommand{\arraystretch}{1.3} 
\setlength{\tabcolsep}{4pt} 
\begin{tabular}{!{\vrule width 1.2pt}c!{\vrule width 1.2pt}c|c!{\vrule width 1.2pt}c|c!{\vrule width 1.2pt}}
\noalign{\hrule height 1.2pt}
\multirow{2}{*}{\textbf{Task}} & \multicolumn{2}{c!{\vrule width 1.2pt}}{\textbf{Quality}} & \multicolumn{2}{c!{\vrule width 1.2pt}}{\textbf{Energy}} \\
\cline{2-5}
& $X^2$ & $p$ & $X^2$ & $p$ \\
\noalign{\hrule height 1.2pt}
De→En & 30.90 & 0.00114 & 25.00 & 0.00912 \\
\hline
En→De & 29.41 & 0.00196 & 31.07 & 0.00107 \\
\hline
GSM8K & 19.99 & 0.0455 & 26.74 & 0.00502 \\
\hline
HumanEval & 26.97 & 0.00464 & 29.74 & 0.00174 \\
\hline
WikiText & 12.36 & 0.337 & 27.67 & 0.00364 \\
\noalign{\hrule height 1.2pt}
\end{tabular}
\vskip -0.1in
\end{table*}

\begin{table*}[h]
\footnotesize
\centering
\caption{Statistical significance (p-values) of Kendall’s Tau correlations between Latency and Energy rankings}
\label{tab:kendall}

\centering
\renewcommand{\arraystretch}{1.3} 
\setlength{\tabcolsep}{4pt} 
\begin{tabular}{!{\vrule width 1.2pt}c!{\vrule width 1.2pt}c|c|c|c|c!{\vrule width 1.2pt}}
\noalign{\hrule height 1.2pt}
\textbf{Model} & De→En & En→De & GSM8K & HumanEval & WikiText \\
\noalign{\hrule height 1.2pt}
Qwen-2.5-7B-Inst & $1.63e^{-05}$ & $3.22e^{-07}$  & $3.22e^{-07}$ & $1.63e^{-05}$ & $1.8e^{-03}$ \\
\hline
Llama-3.1-8B-Inst & $4.99e^{-04}$ & $3.22e^{-07}$ & $3.22e^{-07}$ & $1.63e^{-05}$ & $1.07e^{-04}$ \\
\hline
Qwen-2.5-14B-Inst & $3.18e^{-03}$ & $2.4e^{-04}$ & $5.32e^{-06}$ & $5.01e^{-08}$ & $1.38e^{-02}$ \\
\noalign{\hrule height 1.2pt}
\end{tabular}
\vskip -0.1in
\end{table*}

\end{document}